\documentclass[journal]{IEEEtran}

\usepackage{amsmath}
\usepackage{amsfonts}
\usepackage{amssymb}

\usepackage{amsthm}
\usepackage{mathtools}
\usepackage{tabularx,booktabs}
\usepackage{graphicx}
\usepackage{subfigure}
\usepackage{enumerate}
\usepackage{float}
\usepackage{url}
\usepackage{verbatim}
\usepackage{cite}
\usepackage{diagbox}
\usepackage{multirow}
\usepackage{makecell}
\usepackage{algorithm}  
\usepackage{algorithmicx}  
\usepackage{algpseudocode} 
\usepackage{pifont}
\usepackage[linkcolor=black,citecolor=black,urlcolor=black,colorlinks=true]{hyperref}

\graphicspath{{figures/}}
\IEEEoverridecommandlockouts

\DeclareMathOperator*{\argmin}{arg\,min}

\author{ Lun Quan$^*$, Longji Yin$^*$, Tingrui Zhang, Mingyang Wang, \\ 
        Ruilin Wang, Sheng Zhong, Xin Zhou, Yanjun Cao, Chao Xu, and Fei Gao
    \thanks{ $^*$Indicates equal contribution.}
    \thanks{Corresponding author: Fei Gao, Chao Xu, and Yanjun Cao.}
    \thanks{This work was supported by the National Natural Science Foundation of China under grant no. 62003299 and 62088101. All authors are from the State Key Laboratory of Industrial Control Technology, Institute of Cyber-Systems and Control, Zhejiang University, Hangzhou 310027, China, and the Huzhou Institute, Zhejiang University, Huzhou 313000, China 
    \{lunquan, fgaoaa, cxu, yanjunhi\}@zju.edu.cn, ljyin6038@163.com.
    }
}

\title{\LARGE \bf Robust and Efficient Trajectory Planning for \\ Formation Flight in Dense Environments}

\begin{document}
    \maketitle

\begin{abstract}
Formation flight has a vast potential for aerial robot swarms in various applications. However, existing methods lack the capability to achieve fully autonomous large-scale formation flight in dense environments. To bridge the gap, we present a complete formation flight system that effectively integrates real-world constraints into aerial formation navigation. 
This paper proposes a differentiable graph-based metric to quantify the overall similarity error between formations. This metric is invariant to rotation, translation, and scaling, providing more freedom for formation coordination. We design a distributed trajectory optimization framework that considers formation similarity, obstacle avoidance, and dynamic feasibility. The optimization is decoupled to make large-scale formation flights computationally feasible. To improve the elasticity of formation navigation in highly constrained scenes, we present a swarm reorganization method that adaptively adjusts the formation parameters and task assignments by generating local navigation goals. 
A novel swarm agreement strategy called global-remap-local-replan and a formation-level path planner is proposed in this work to coordinate the global planning and local trajectory optimizations.
To validate the proposed method, we design comprehensive benchmarks and simulations with other cutting-edge works in terms of adaptability, predictability, elasticity, resilience, and efficiency.
Finally, integrated with palm-sized swarm platforms with onboard computers and sensors, the proposed method demonstrates its efficiency and robustness by achieving the largest scale formation flight in dense outdoor environments.
\end{abstract}


\vspace{-0.5cm}
\section{Introduction}
\label{sec:intro}
\IEEEPARstart{F}{ormation} flight has become a fundamental capability for autonomous swarms to achieve coordinated aerial maneuvers. In cluttered wilds and complex urban areas, formation navigation has a wide potential in search and rescue\cite{marconi2012sherpa}, collaborative mapping\cite{mahdoui2020communicating}, package delivery\cite{dorling2016vehicle}, and so on. 
However, effectively integrating real-world constraints into an aerial formation remains an unresolved problem. 
This article aims to empower aerial swarms to maintain cooperative formation behaviors in dense environments by proposing a complete formation flight system.

Inspired by natural swarm systems like bird flocks and fish schools, an ideal formation flight system should possess the capability to flexibly adapt and deform in dense environments.
By striving to maintain the swarm in a ``critical state'', swarm formation can dynamically balance the conflicts between maintaining formation and avoiding obstacles.

\begin{figure}
    \begin{center}
        \includegraphics[width=0.9\columnwidth]{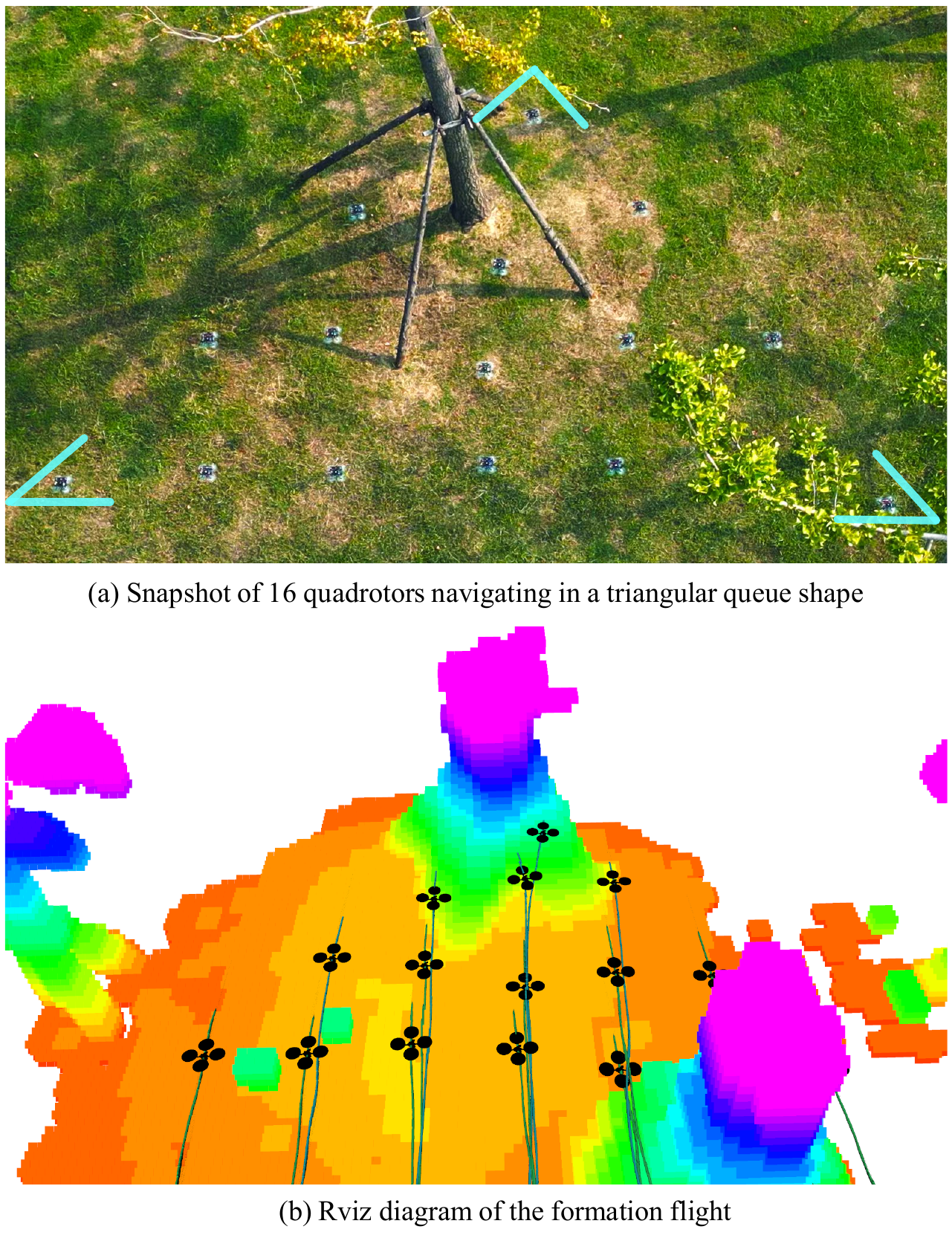}
    \end{center}
    \vspace{-0.5cm}
    \caption{Large-scale formation flight in the dense outdoor environment. (a) Snapshot of the moment the swarm robots prepare to fly through the woods. (b) Rviz diagram of the executed trajectories. The grid map is merged using log data offline. Please watch our attached videos for more information about the experiments at \url{https://www.youtube.com/watch?v=uEMyvPxYqmA}.}
    \label{fig:main}
    \vspace{-0.5cm}
\end{figure}

While extensive research works focus on navigation in formation, few achieve robust formation flights in obstacle-rich areas. 
Three core challenges limit practical formation applications:
(a) The inherent conflict between formation maintenance and obstacle avoidance is inevitable and difficult to mitigate.
(b) Predefined formations lack elastic adaptability in response to constrained environments.
(c) The swarm system cannot rapidly recover from disordered states caused by unknown obstacles or sudden changes in the desired formation shape.

Based on the above challenges, we conclude that an ideal formation flight system should have the ability to maintain formation while avoiding obstacles, adjust swarm formation distributions according to constrained environments, and reorganize the formation quickly after emergencies.
These characteristics are summarized as the \textit{PAPER} criteria:
\begin{itemize}
    \item \textbf{P}ortability: Aerial robotic swarms should comprise lightweight platforms with scalable systems and distributed architecture.
    A scalable system means the main components, such as estimation, decision, planning, and control modules, are all the same on each robot. 
    A distributed architecture is inherently robust against individual hardware failures.
    These are the basis for large-scale formation flight.
    \item \textbf{A}daptability: When facing obstacles, robots should locally adapt their trajectories to avoid collision in a way that does the least damage to the overall formation performance. This ability mitigates the conflict between formation maintenance and obstacle avoidance.
    \item \textbf{P}redictability: Reactive local feedback methods are short-sighted and can not consider the constraints in advance. Robots should optimize the motions over a prediction horizon so that the formation can respond smoothly to the future environmental changes in its vicinity, which is necessary for dense areas.
    \item \textbf{E}lasticity: A feasible and safe trajectory for a fixed formation shape may not exist in constrained environments, such as narrow corridors or holes. Therefore, swarm robots need to have elastic and flexible deformation capabilities by adjusting formation distributions (such as the scale of shape or task assignments) while keeping the full maneuverability of the formation.
    \item \textbf{R}esilience: Formation flight could encounter many unfavorable situations caused by unknown obstacles or sudden changes of desired formation shape. The navigation system should be able to resiliently reorganize and guide the whole formation so that the flight can recover from disordered states timely.
\end{itemize}
A complete formation flight system should meet the above \textit{PAPER} criteria and ensure that the conditions for each criterion are compatible with the others.

Our previous work\cite{quan2022formation} only partially met the first three terms of \textit{PAPER} criteria.
We tackled formation flight as a coupled collaborative trajectory optimization problem, suitable primarily for small-scale formation scenarios.
However, resolving cooperative constraints of the formation using the graph-based similarity metric was computationally heavy, resulting in increased overhead during each optimization iteration.
Moreover, integrating dynamic inter-robot relationships within the coupled trajectory optimization problem considerably affected the efficiency of the optimization process, rendering it less appropriate for larger formations or more complex scenarios.

In this paper, we present a complete formation flight system that satisfies all \textit{PAPER} criteria.
To address the challenges in [4], we introduce a decoupled formation optimization method to significantly improve computational efficiency.
This method consists of two components.
Firstly, an optimal formation position sequence is pre-computed, avoiding repetitive metric calculation during the optimization process.
Secondly, a fixed time interval sampling method is used to convert dynamic inter-robot relationships into static constraints, greatly reducing the complexity of the optimization problem.
These improvements make our method suitable for large-scale swarms.
Besides, the previous method lacks the capability of reorganizing the swarm formation, which may lead to disordered formation flights under adverse conditions, especially when the initial positions or task assignments are inappropriate.
To address this, we propose a swarm reorganization method that can elastically adjust formation distributions by optimizing formation parameters and task assignments in response to external constraints.
Subsequently, we develop a swarm agreement strategy called global-remap-local-replan, which enables rapid implementation of the swarm reorganization results to achieve consensus among swarm agents.
Additionally, a formation-level global path-finding method, which treats the swarm formation as one entity, is also designed to guide the swarm out of the obstacle deadlocks.
Finally, we integrate the estimation, mapping, decision, planning, and control modules into palm-sized swarm platforms \cite{zhou2022swarm} with onboard computers and sensors, enabling large-scale formation flight in dense environments.
Detailed contributions are as follows.
\begin{enumerate}
    \item We introduce an optimal formation position sequence, pre-computed using the differentiable graph-based metric~\cite{quan2022formation}. This sequence represents the optimal position with the lowest similarity error, reducing the need for repetitive computation during the optimization process.
    \item We design a decoupled spatial-temporal trajectory optimization framework that effectively handles dynamic inter-robot relationships, obstacle avoidance, and dynamic feasibility.
    Compared to our prior work~\cite{quan2022formation}, we achieve higher computational efficiency for large-scale swarms.
    \item We present a swarm reorganization method to achieve elastic deformation of swarm distributions, which simultaneously solves optimal formation alignment and task assignment problems (ALAS for short). This method improves the elasticity of swarm formation against constrained environments. 
    It relieves the dependence on the appropriate formation alignments and task assignments.
    \item We design a global-remap-local-replan strategy (GRLR for short) that leverages the advantages of centralized formation parameter remapping and decentralized local trajectory replanning.
    With this strategy, the distributed asynchronous swarm is able to quickly recover from disordered states and return to formation flight quickly.
    \item We integrate all these modules into a hierarchical formation flight system.
    Extensive benchmarks and simulations are conducted to validate the \textit{PAPER} criteria of our method.
    A series of real-world experiments are designed to demonstrate the outstanding performance of the proposed distributed autonomous formation flight system.
\end{enumerate}

\section{Related works}
\label{sec: related works}
\subsection{Distributed Swarm Trajectory Planning}
\label{subsec: swarm planninng}
Extensive works exist for trajectory planning of distributed swarms. 
The concept of velocity obstacle (VO) is leveraged and generalized by Van Den Berg et al.\cite{van2011RR,van2011reciprocal,bareiss2013reciprocal} to accomplish reciprocal collision avoidance for multiple robots. 
However, the smoothness of the resulting trajectories cannot be guaranteed by VO-based approaches, which significantly impairs the usability of the actual robot systems.

In order to produce high-quality collision-free trajectories, optimization-based methods are widely introduced in the literature on distributed multicopter swarms\cite{arul2020dcad,luis2019trajectory,park2020efficient}. Zhou et al.\cite{zhou2017fast} incorporate Voronoi cell tessellation into a receding horizon QP scheme to prevent collision among the robots while planning. 
In \cite{chen2015decoupled}, Chen et al. employ SCP to address the multiagent planning problem in non-convex space by incrementally tightening the collision constraints. Baca et al.\cite{baca2018model} combine MPC with a conflict resolution strategy to ensure mutual collision avoidance for outdoor swarm operations. Nevertheless, the computational load of the above optimization-based methods is large, which could hamper the applicability of the planners in highly dense scenarios.   

Recently, Zhou et al.\cite{zhou2022swarm} present a distributed autonomous quadrotor swarm system using spatial-temporal trajectory optimization, which generates collision-free motions in dense environments merely in milliseconds. Our distributed formation trajectory optimization is based on this work.
\vspace{-0.5cm}

\subsection{Formation Flight in Free Space}
\label{subsec: formation flight in free space}
Various techniques have been proposed to achieve multi-robot navigation in formation, which include virtual structures\cite{lewis1997high}, leader-follower\cite{panagou2014cooperative}, navigation functions\cite{de2006formation}, reactive behaviors\cite{balch1998behavior}, consensus-based local control laws\cite{lin2013leader}, and barycentric-coordinate-based control \cite{fathian2020robust}. However, most of the existing methods only consider obstacle-free cases.

Weinstein et al.\cite{weinstein2018visual} present a VIO-swarm system that performs all modules onboard and can execute formation flight without inter-robot collisions in free space.
Parker et al.\cite{parker2018pipeline} present a distributed formation control method and relax the dependency of the common reference frame.

As the scale of swarms increases, researchers begin to notice that it is difficult to maintain the formation only by trajectory planning, especially when there are deadlocks between robots.
Turpin et al.\cite{turpin2014capt} consider the problem of concurrent assignment and collision-free trajectory generation. 
Turpin gives centralized and decentralized solutions to this problem, allowing flight formation on a large scale.
Morgan et al.\cite{morgan2016swarm} also use model predictive control to solve task assignment and trajectory generation simultaneously when given the desired formation shape.
In addition to considering task assignments, Agarwal and Akella\cite{agarwal2018simultaneous} consider formation alignment problems to optimize the formation parameters such as scale and location.
This method reduces the cost of forming formation and speeds up convergence.
However, these methods ignore the influence of constrained environments, in which formation should elastically deform to navigate.

\vspace{-0.5cm}
\subsection{Formation Flight in constrained Environments}
\label{subsec: formation flight}
In constrained environments, where various obstacles and limitations exist, formation flight can be a challenging task that requires constant adjustments to maintain the swarm structure.
An immediate solution is to design composite control laws that combine formation flight and collision avoidance by using multiple layered potential fields\cite{zhou2018agile}, which are prone to deadlock.
A better solution is to allow the formation shape to deform while maintaining the overall swarm structure.
Han et al. \cite{han2013local} propose a complex-valued graph Laplacian-based formation controller that regulates the scaling of formation shape during swarm maneuvering like passing through corridors. 
In \cite{zhao2018affine}, Zhao proposes a leader-follower control law enabling the affine transformation of formation in response to environmental changes.
And the bearing-based local controller \cite{zhao2019baering, zhao2019bearingLong} exhibits translational, scaling, and rotational invariance of formation flight.
However, these methods rely on leaders or predefined trajectories and struggle with complex obstacles or sudden potential collisions.

Compared to the local feedback methods, predictive optimization-based methods proactively plan the future motion of swarm robots, striking a balance between formation flight and obstacle avoidance.
Alonso-Mora et al.\cite{alonso2016distributed} control swarm robots by optimally rearranging the desired formation and planning local trajectories for each drone. 
However, since there is no inter-vehicle coordination in the distributed planners, formation maintenance is not conducted during local planning.
Peng et al.\cite{peng2022obstacle} propose a method to improve flight safety by enabling the affine transformation of formation shape and treating it as a soft constraint during B-spline optimization.
However, this approach requires optimizing the trajectories of all robots simultaneously and cannot be applied to large-scale swarms.
To tackle formation preservation, Parys et al.\cite{van2017distributed} propose a distributed model predictive formation controller. 
This framework imposes relative position constraints on the swarm and coordinates the agents to break passively once obstacles violate positional constraints.
Overall, these approaches offer unique solutions for trajectory planning in swarm robotics, but they each have limitations when dealing with different scenarios and scales of robotic systems.

To address these drawbacks, we formulate the overall formation requirement with a differentiable metric in trajectory optimization.
This allows us to fully utilize the collaboration ability of the swarm, effectively avoid deadlock, and foresee obstacle avoidance.
Besides, we adopt a distributed and decoupled optimization method to ensure dynamic real-time performance.
This approach can be applied to large-scale swarms while still maintaining efficient trajectory planning.

\begin{figure*}
    \begin{center}
        \includegraphics[width=1.8\columnwidth]{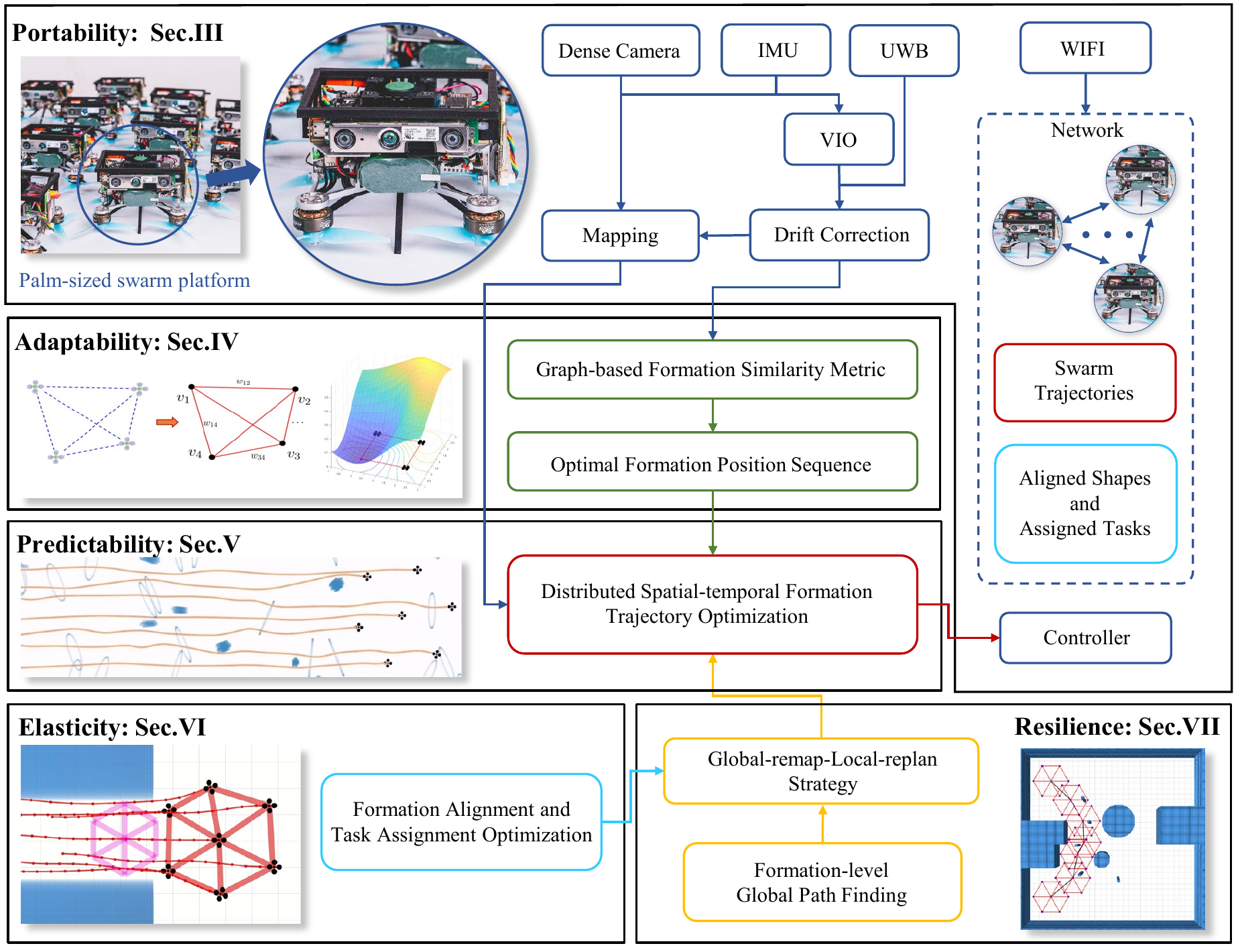}
    \end{center}
    \vspace{-0.5cm}
    \caption{Illustration of the system architecture. In order to facilitate understanding, we divide the various modules of the formation flight system according to the \textit{PAPER} criteria mentioned in Sec.\ref{sec:intro}. The main challenge in integrating \textit{PAPER} criteria into a swarm formation flight system is to ensure that the conditions for each criterion are compatible with the others. Therefore we build a hierarchical formation flight system. We use different colors to represent different levels in the system. Moreover, the information broadcast through the network is from the same color module.}
    \label{fig:system}
    \vspace{-0.5cm}
\end{figure*}

\section{System Overview}
\label{sec:system}
This paper aims to optimize the autonomy of swarm robots in real-world environments by coordinating their movements to form a desired formation shape. 
To accomplish this, we adopt a distributed swarm aerial robot system and propose a spatial-temporal trajectory optimization for formation flight. 
To enhance the system's robustness, we also address the case of swarm disorder by incorporating an adaptive swarm reorganization method and an efficient swarm agreement strategy.

\vspace{-0.5cm}
\subsection{Swarm Aerial Robot System}
\label{subsec:platform2}
The swarm aerial robot system is composed of palm-sized quadrotor platforms\cite{zhou2022swarm} with depth stereo camera\footnote{https://www.intelrealsense.com/depth-camera-d435/} for imagery and depth sensing, as shown in Fig.~\ref{fig:system}. 
The software modules, including state estimation, environment perception, decision-making, trajectory planning, and flight control, run in real-time on an onboard computer\footnote{https://www.nvidia.com/en-us/autonomous-machines/embedded-systems/jetson-xavier-nx/}. 
This lightweight and scalable platform suit dense environments.

We use visual-inertial odometry (VIO)\cite{qin2017vins} to estimate each robot's pose with respect to its start frame, and we recover the transformations related to the start frames with only anonymous bearing measurements in our previous work\cite{wang2022Ral}. 
For simplicity, the transformations are known in our experiments by requiring robots to take off from pre-defined locations. 
To correct the localization drift between swarm robots, we use a drift correction method\cite{zhou2022swarm} with onboard ultra-wideband (UWB).

The distributed system architecture enables each robot to fully utilize its computing resources to process more information, relieving the pressure of network communication. 
The robots share only important information, such as trajectories, for high-fidelity wireless communication, and there is no ground station to send control inputs.

\vspace{-0.5cm}
\subsection{Distributed Local Formation Trajectory Optimization}
\label{subsec:distributed local trajectory}
The local formation trajectory optimization is distributed and asynchronous, which enables each robot to generate its trajectory only depending on local information and does not require the same start timestamp or same time duration of trajectory.
Each robot evaluates the formation state by calculating the formation similarity error and generates its optimal formation position sequence (Sec.\ref{sec:formation}) to maintain the desired formation shape.
Then a subsequent trajectory optimization module generates spatial-temporal formation trajectories (Sec.\ref{sec:trajectory optimization}) for real-time navigation.
The above process cycles periodically within the receding horizon. 
The distributed local formation trajectory optimization can always maintain the overall formation when navigating in a complex environment.

\vspace{-0.5cm}
\subsection{Swarm Reorganization and Agreement Methods}
\label{subsec:consensus method}
Only relying on the distributed local trajectory optimization may lead to poor formation flight quality when encountering narrow corridors or instant transformation of formation shapes.
Therefore, we propose swarm reorganization (Sec.\ref{sec:swarm consensus}) and agreement methods (Sec.\ref{sec:swarm agree}) to achieve the swarm consensus quickly.
Inspired by the flock flying behavior of birds, we design a global-remap-local-replan (GRLR for short) strategy, which only centrally remaps crucial parameters of swarm formation and makes the swarm formation converge quickly through distributed replanning local trajectory.
Firstly, when the stable state of the swarm formation is destroyed or about to be destroyed, the swarm robots designate a leader (drone 0 in this paper).
The crucial parameters of swarm formation are calculated separately by formation alignment and task assignment optimization (ALAS for short) and formation-level global pathfinding method.
From the perspective of swarm reorganization and agreement, the conflict between swarm formation and obstacle is alleviated by optimizing formation alignment, and the speed of formation convergence is greatly accelerated by optimizing task assignment.
GRLR strategy is simple but very effective by combining distributed methods' efficiency and centralized ones' optimality.

\section{adaptive description of swarm formation}
\label{sec:formation}
\subsection{Graph-based Formation Definition}
\label{subsec:graph}
In this paper, a swarm formation of $N$ robots is modeled by an undirected graph $\mathcal{G} = (\mathcal{V,E})$, where $\mathcal{V}:=\{1,2,...,N\}$ is the set of vertices, and $\mathcal{E} \subset \mathcal{V} \times \mathcal{V}$ is the set of edges. In graph $\mathcal{G}$, the vertex $i$ represents the $i^{th}$ robot with position vector $\mathbf{p}_i = [x_i,y_i,z_i] \in \mathbb{R}^3$ . An edge $e_{ij} \in \mathcal{E}$ that connects vertex $i\in \mathcal{V}$ and vertex $j\in \mathcal{V}$ means that robot $i$ and $j$ can measure the geometric distance between each other. 
In our work, each robot can obtain the positions of all robots $\{\mathbf{p}_1,...,\mathbf{p}_i,...,\mathbf{p}_N\}$, thus the graph $\mathcal{G}$ is complete.
Then we determine the adjacency matrix $\mathbf{A} \in \mathbb{R}^{N\times N}$ and degree matrix $\mathbf{D}\in \mathbb{R}^{N\times N}$ of the formation graph $\mathcal{G}$ by
\begin{equation}
\label{equ:A_weight}
    A_{ij}=w_{ij}= \parallel \mathbf{p}_i-\mathbf{p}_j\parallel^2,
\end{equation}
\begin{equation}
\label{equ:D_weight}
    D_{ij}=\begin{cases}
                \sum_{j = 1}^{N} A_{ij},  & \text{if}~~i=j, \\
                0, & \text{otherwise},
            \end{cases}
\end{equation}
where the non-negative edge weight $w_{ij}$ is the squared distance between the $i^{th}$ and $j^{th}$ robots, and $\parallel\cdot\parallel$ denotes the Euclidean norm.
Thus, the corresponding Laplacian matrix is
\begin{equation}
\label{equ:L}
    \mathbf{L} = \mathbf{D} - \mathbf{A}.
\end{equation}

With the above matrices, the symmetric normalized Laplacian matrix of graph $\mathcal{G}$ is defined as
\begin{equation}
\label{equ:L_normalization}
    \mathbf{\hat{L}} = \mathbf{D}^{-1/2}\mathbf{L}\mathbf{D}^{-1/2} = \mathbf{I} - \mathbf{D}^{-1/2}\mathbf{A}\mathbf{D}^{-1/2},
\end{equation}
where $\mathbf{I} \in \mathbb{R}^{N\times N}$ is the identity matrix.
$\mathbf{\hat{L}}$ contains the information that is invariant to scale, translation, and rotation.

Finally, we use graph theory to describe various desired formation shapes,  such as squares, hexagons, and pyramids.
By specifying the positions $\mathbf{p}_i^d= [x_i^d,\,\,y_i^d,\,\,z_i^d]\in \mathbb{R}^3,\,\,i = 1,...,N$, computing $\mathbf{\hat{L}}_{des}$ is simple.
It's important to note that the desired formation shape is independent of the coordinate system as long as the relative positions are provided.

\vspace{-0.5cm}
\subsection{Differentiable Formation Similarity Error Metric}

\begin{figure}
    \begin{center}
         \includegraphics[width=1.0\columnwidth]{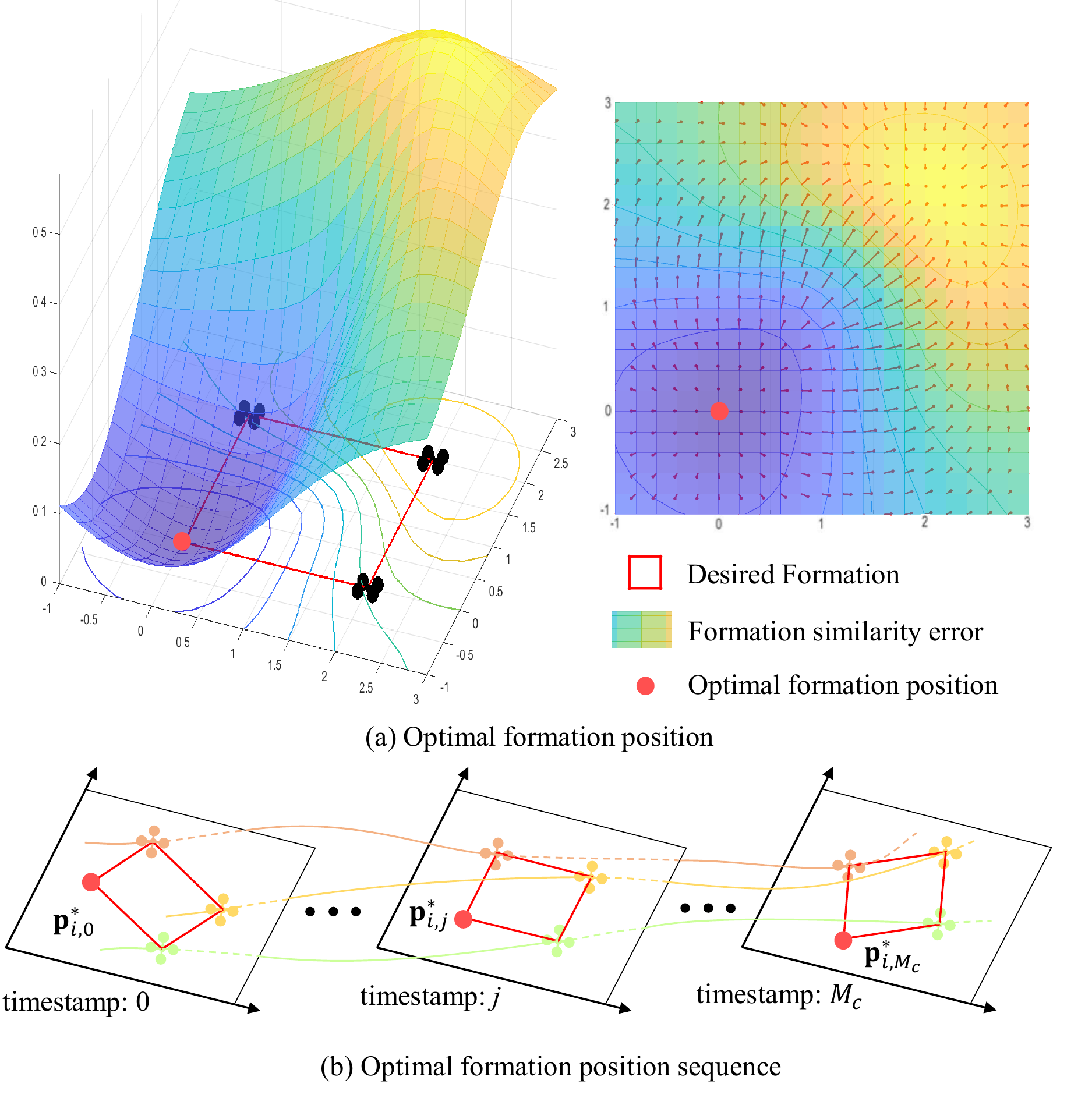}
    \end{center}
    \vspace{-0.5cm}
    \caption{Illustration of optimal formation position sequence using a 2D formation. (a) The surface shows the profile of the similarity metric when one UAV moves in the plane and the other three remain still. The minimum suggests the optimal formation position to form the desired shape. (b) The sequence of optimal formation positions corresponds to the timestamps.}
    \label{fig:optimal_formation_sequence}
    \vspace{-0.5cm}
\end{figure}

To assess the deviation from the desired formation, we propose a differentiable formation similarity error metric as
\begin{equation}
\label{equ:L_metric}
    \begin{aligned}
        f_s & = f_s(\mathbf{p}_1,...,\mathbf{p}_i,...,\mathbf{p}_N) = f_s (\mathbf{A},\mathbf{D}) = f_s(\mathbf{\hat{L}},\mathbf{\hat{L}}_{des}) \\   
        & = \parallel\mathbf{\hat{L}}-\mathbf{\hat{L}}_{des}\parallel^2_F = tr\{(\mathbf{\hat{L}}-\mathbf{\hat{L}}_{des})^T(\mathbf{\hat{L}}-\mathbf{\hat{L}}_{des})\},
    \end{aligned}
\end{equation}
where $tr\{\cdot\}$ denotes the trace of a matrix, $\mathbf{\hat{L}}$ is the symmetric normalized Laplacian of the current swarm formation, $\mathbf{\hat{L}}_{des}$ is the counterpart of the desired formation. Frobenius norm $\parallel\cdot\parallel_F$ is used in our distance metric. 
As a graph representation matrix, $\mathbf{\hat{L}}$ contains information about the graph structure\cite{NatureReport}.
This allows $f_s$ to consider only the geometric shape of the formation, and not be influenced by scaling, translation, or rotation.
Additionally, $f_s$ is a dimensionless value that solely reflects the error in formation shape similarity.

In particular, under the distributed framework, each robot can only change its positions to reduce the overall formation similarity error. 
Therefore, the only variable for robot $i$ in (\ref{equ:L_metric}) is $\mathbf{p}_i$, and $f_s(\mathbf{p}_1,...,\mathbf{p}_i,...,\mathbf{p}_N)$ can be simplified as $f_s(\mathbf{p}_i)$.
    
Our metric is analytically differentiable with respect to the position of each robot. For robot $i$, we use the weights of its $N$ adjacent edges $\{e_{i1},\,e_{i2},\,...,\;,e_{iN}\}$ to form a weight vector $\mathbf{w}_{i} = [w_{i1},\,w_{i2},\,...,\;,w_{iN}]^T$. By the chain rule, the gradient of $f_s$ with respect to $\mathbf{p}_i$ can be written as 
\begin{equation}
    \label{eq:f_p}
    \frac{\partial f_s}{\partial \mathbf{p}_i} = \frac{\partial f_s}{\partial \mathbf{w}_i^T} \frac{\partial \mathbf{w}_i}{\partial \mathbf{p}_i}.
\end{equation}
According to our metric (\ref{equ:L_metric}), the gradient of $f_s$ with respect to each weight $w_{ij}$ can be computed as follow
\begin{equation}
\begin{aligned}
    \frac{\partial f_s}{\partial w_{ij}} &= tr\{(\frac{\partial f_s}{\partial \hat{\mathbf{L}}})^T(\frac{\partial \hat{\mathbf{L}}}{\partial w_{ij}})\}, \\
    \frac{\partial f_s}{\partial \mathbf{\hat{L}}} &= \frac{\partial ||\mathbf{\hat{L}}-\mathbf{\hat{L}}_{des}||_F^2}{\partial \mathbf{\hat{L}}} =
2(\mathbf{\hat{L}}-\mathbf{\hat{L}}_{des}), \\
    \frac{\partial \mathbf{\hat{L}}}{\partial w_{ij}} &=
    -\frac{\partial(\mathbf{D}^{-1/2}\mathbf{A}\mathbf{D}^{-1/2})}{\partial w_{ij}}.
\end{aligned}
\end{equation}
Then the gradient $\partial f_s/\partial \mathbf{w}_i$ can be written as 
\begin{equation}
   \partial f_s/\partial \mathbf{w}_i = [\partial f_s/\partial w_{i1}, \partial f_s/\partial w_{i2}, ..., \partial f_s/\partial w_{iN}]^T.
\end{equation}
As for $\partial \mathbf{w}_i/\partial \mathbf{p}_i$, the Jacobian can be easily derived since the weight function (\ref{equ:A_weight}) is a differentiable quadratic form.

\vspace{-0.5cm}
\subsection{Optimal Formation Position Sequence}
\label{subsec:optimization formation}
In our previous work \cite{quan2022formation}, we incorporated $f_s$ directly into the trajectory optimization, making formation flight a coupled trajectory optimization problem.
While this method is suitable for small-scale formation flight, it becomes computationally inefficient as the number $N$ of robots increases.
Considering the simplified equation for coupled trajectory optimization
\begin{equation}
    \label{equ:coupled}
    \min_{\mathbf{p}_{i,0},...,\mathbf{p}_{i,M_c}} \sum_{j=0}^{M_c} f_s(\mathbf{p}_{i,j})+J_{other},
\end{equation}
where $\mathbf{p}_{i,j}$ represent the $j^{th}$ sample point of $i^{th}$ robot trajectory in (\ref{equ:sampling points}) for convenience.
$J_{other}$ represents all other cost functions, and $M_c$ is the number of sample points with corresponding timestamps.
The primary purpose of calculating $f_s$ is to supply gradient information for minimizing formation similarity error.
However, since the graph $\mathcal{G}$ is a complete graph, computing $f_s$ has a complexity of $O(N^2)$.
Consequently, the coupled trajectory optimization (\ref{equ:coupled}) also exhibits high complexity of $O(N^2)$ in each iteration, limiting its applicability to large-scale swarm operations.

To address this issue, we must identify an equivalent approach with reduced computational complexity to replace the function of $f_s$ in (\ref{equ:coupled}).
We introduce the concept of \textbf{optimal formation position} $\mathbf{p}_{i,j}^*$ for robot $i$ at timestamp $j$, which is the position that minimizes the formation similarity error $f_s$.
Fig.~\ref{fig:optimal_formation_sequence}(a) illustrates this concept using a 2D formation as an example. 
It is evident from the figure that there exists an optimal formation position $\mathbf{p}_{i,j}^*$ that results in a minimal formation similarity error, and the partial derivative is $\partial f_s/\partial \mathbf{p}_{i,j}^*=0$.
In the future period with a sequence of timestamps $\{0, ..., j, ..., M_c\}$, we represent the expected positions of robot $i$ with the \textbf{optimal formation position sequence} $\mathbf{p}_i^*=\{\mathbf{p}_{i,0}^*,\cdots,\mathbf{p}_{i,j}^*,\dots,\mathbf{p}_{i,M_c}^*\}$, as shown in Fig.~\ref{fig:optimal_formation_sequence}(b).
By precomputing $\mathbf{p}_i^*$, we can utilize its quadratic distance to replace the gradient information offered by $f_s$ in (11), thus decreasing the computational requirements as follows
\begin{equation}
    \label{equ:fs_to_p}
    f_s(\mathbf{p}_{i.j}) \Rightarrow \| \mathbf{p}_{i,j} - \mathbf{p}_{i.j}^* \|^2.
\end{equation}
Since the optimal solutions of $f_s$ and quadratic distance cost are equivalent, the trajectory approaches the positions with minimal formation similarity error, maintaining the desired formation.
Thus, we can effectively solve the coupled trajectory optimization with a two-step procedure
\begin{equation}
    \label{equ:decoupled}
    \begin{aligned}
         &\text{\ding{172}}~~\mathbf{p}_i^* = \argmin \sum_{j=0}^{M_c}f_s(\mathbf{p}_{i.j}), \\
         \xRightarrow{\mathbf{p}_i^*}~~&\text{\ding{173}}~~\min_{\mathbf{p}_{i,0},...,\mathbf{p}_{i,M_c}} \| \mathbf{p}_{i.j}-\mathbf{p}_{i.j}^*\|^2+J_{other}.
    \end{aligned}
\end{equation}
As a result, the previously required calculation of $f_s$ in each trajectory optimization process is replaced by the computation of the quadratic distance, simplifying the optimization problem.
This significantly reduces computational demands and enables large-scale swarm formation.

Formula (\ref{equ:decoupled}) indicates that trajectory optimization in Sec.\ref{sec:trajectory optimization} is performed on discretized points. 
Non-uniform discretized points may lead to poor trajectories and sub-optimal performance.
Therefore it is crucial to ensure a uniform distribution of these points to maintain the effectiveness of the optimization process.
In engineering practice, since graphs $\mathcal{G}$ are constructed from a series of discretized timestamps as depicted in Fig.~\ref{fig:optimal_formation_sequence}(b), each $\mathbf{p}_{i,j}^*$ is independent.

To ensure a smoother trajectory, we introduce the uniform optimal formation position sequence $\hat{\mathbf{p}}_i^*$, which is generated by considering the formation similarity error $J_s$ and the uniform distribution cost $J_u$
\begin{equation}
    \label{equ:formation position problem}
        \hat{\mathbf{p}}_i^* = \argmin \lambda_s J_s + \lambda_u J_u,
\end{equation}
\begin{equation}
\label{equ:J_u_}
    \begin{aligned}
        J_s &= \sum_{j=0}^{M_c} 
        f_s(\hat{\mathbf{p}}_{i,j}^*), \\
        J_u &= \mathbb{E}(\mathbf{U}^2) - \mathbb{E}(\mathbf{U})^2 = \frac{\| \mathbf{U}\|_2^2}{M_c}  - \frac{\| \mathbf{U}\|_1^2}{(M_c)^2},
    \end{aligned}
\end{equation}
where $\lambda_s$ and $\lambda_u$ are the relative weights.
$\mathbb{E}(\cdot)$ is mathematic expectation and the squared distance vector $\mathbf{U} \in \mathbb{R}^{M_c}$ is 
\begin{equation}
\label{equ:U_vector}
    \mathbf{U}=(\|\hat{\mathbf{p}}_{i,1}^*-\hat{\mathbf{p}}_{i,0}^*\|_2^2,\cdots,\|\hat{\mathbf{p}}_{i,M_c}^*-\hat{\mathbf{p}}_{i,M_c-1}^*\|_2^2).
\end{equation}

We use the quasi-Newton method \cite{lewis2013nonsmooth} to solve this unconstrained optimization problem (\ref{equ:formation position problem}) and generate uniform $\hat{\mathbf{p}}_i^*$ for the later trajectory optimization (\ref{equ:unconstainted problem}).
By doing so, the trajectory resulting from these discretized points in Sec.\ref{sec:trajectory optimization} can be smoother and avoid sudden spatial changes.

\section{Spatial-temporal trajectory optimization for formation flight}
\label{sec:trajectory optimization}
\subsection{Trajectory Representation}
\label{subsec:trajectory}
The differential flatness of multicopters \cite{MelKum1105} benefits trajectory generation without integrating differential equations. 
Moreover, the motion planning of multicopters can be performed on low-dimensional smooth trajectories.
In this paper, we adopt a state-of-the-art trajectory representation named MINCO~\cite{wang2021geometrically} to achieve minimum control effort spatial-temporal trajectory planning for swarm aerial robots in three-dimensional environments. 
MINCO conducts spatial-temporal deformation of the flat-output $M$-piece trajectory $p(t)$ by decoupling the space and time parameters with a linear-complexity mapping $\mathcal{M}$
\begin{equation}
    \label{equ:mapping}
    p(t)=\mathcal{M}_{\mathbf{q},\mathbf{T}}(t),~~\forall{t}\in[t_0,t_M],
\end{equation}
where $\mathbf{q}=(\mathbf{q}_1,\cdots,\mathbf{q}_{M-1})^T\in \mathbb{R}^{3\times(M-1)}$ are the adjacent intermediate points between each pair of connected pieces and $\mathbf{T}=(T_1,\cdots,T_M)^T\in \mathbb{R}^M_{>0}$ the time duration of each piece.

A $m$-dimensional $M$-piece trajectory $p(t)$ is represented by piecewise polynomials. And $i^{th}$ piece $p_i(t)$ is defined as a multi-degree polynomial ($Q=5$ in this paper)
\begin{equation}
	\label{equ:i-th piece}
	p_i(t)=\mathbf{c}_i^T \boldsymbol{\beta}(t),~~\forall{t}\in[0,T_i],
\end{equation}
where $\mathbf{c}_i\in \mathbb{R}^{(Q+1)\times m}$ is the coefficient matrix and $\boldsymbol{\beta}(t)=[t^0,t^1,\cdots,t^Q]^T$ is the natural basis.

For an s-integrator ($s=3$ in this paper) chain dynamics system, a $M$-piece $2s-1$ degree trajectory $p(t)$ is defined by constant boundaries and minimum control effort $\{\mathbf{q},\mathbf{T}\}$.
Furthermore, MINCO is advanced in convert $\{\mathbf{q},\mathbf{T}\}$ to $\{\mathbf{c},\mathbf{T}\}$ using a linear-time and space parameter mapping $\mathbf{c}=\mathcal{M}(\mathbf{q},\mathbf{T})$, where $\mathbf{c}=(\mathbf{c}_1^T,\cdots,\mathbf{c}_M^T)^T$ is polynomial coefficients.

\vspace{-0.5cm}
\subsection{Problem Formulation}
After determining the desired formation shape in Sec.\ref{sec:formation}, we expect a cluster of trajectories for swarm robots, which are smooth, collision-free, and formation maintained. 
In practice, navigating swarm robots in an unknown dense environment with FOV-limited sensors and onboard computer requires an efficient real-time planner focusing on local information.
Besides, centralized optimization is limited by the scale of the swarm.
Therefore, we choose a distributed local trajectory optimization for formation flight as follows
\begin{subequations}
\label{equ:local trajectory problem}
    \begin{align}
        \min_{
                \begin{scriptsize}
                \mathbf{q},\mathbf{T}
                \end{scriptsize}
                } ~~
        & \label{equ_local:cost} \int_{t_0}^{t_M} \| p^{(s)}(t) \|^2 dt + \rho \cdot T_\Sigma ,\\
s.t.~~~~& \label{equ_local:system} p(t) = \mathcal{M}_{\mathbf{q},\mathbf{T}}(t), \forall t \in [t_0,t_M], \\
        & \label{equ_local:boundary_1} \mathbf{p}^{[s-1]}(0)=\bar{\mathbf{p}}_{0}, \\
        & \label{equ_local:boundary_2} \mathbf{p}^{[s-1]}(t_M)=\bar{\mathbf{p}}_{f}, \\
        & \label{equ_local:inequal} \mathcal{H}(p(t),...,p^{(s)}(t))\preceq \mathbf{0}, \forall t \in [t_0,t_M].
    \end{align}
\end{subequations}
We define costs (\ref{equ_local:cost}) for smoothness and aggressiveness to achieve smooth and efficient flight.
$\rho$ is time regularization parameter, $T_\Sigma=\sum_{i=1}^M T_i$.
The state of robot $p(t)$ (\ref{equ_local:system}) is parameterized by the optimization variables $\{\mathbf{q},\mathbf{T}\}$.
$\mathbf{p}^{[s-1]}(t)=(p(t)^T,\dot{p}(t)^T,...,p^{(s-1)}(t)^T)^T\in\mathbb{R}^{ms}$ represents the higher-order derivatives of a chain dynamic system with $s$-integrator.
Boundary conditions involve initial state $\bar{\mathbf{p}}_0\in\mathbb{R}^{ms}$ (\ref{equ_local:boundary_1}) and terminal state $\bar{\mathbf{p}}_f\in\mathbb{R}^{ms}$ (\ref{equ_local:boundary_2}).
Continuous-time constraints $\mathcal{H}$ (\ref{equ_local:inequal}) include swarm formation similarity, dynamic feasibility, obstacle avoidance, and swarm reciprocal avoidance. 

\vspace{-0.5cm}
\subsection{Constraints Transcription}
\label{sec:Constraints transcription}
To solve the continuous constrained optimization problem (\ref{equ:local trajectory problem}) in real-time, we use the optimization variable of MINCO (\ref{equ:mapping}) to eliminate all kinds of equality constraints (\ref{equ_local:system})-(\ref{equ_local:boundary_2}).
And penalty function method \cite{jennings1990computational} is used to deal with the inequality constraints (\ref{equ_local:inequal}).
Then, every integral is evaluated by a finite sum of sample points.
Finally, the continuous constrained optimization problem is converted to a discrete unconstrained optimization problem
\begin{equation}
\label{equ:unconstainted problem}
    \begin{aligned}
        \min_{
              \begin{scriptsize}
                \mathbf{q},\mathbf{T}
              \end{scriptsize}
             }  
        \sum_{x} \lambda_\star \widetilde{J}_\star(\mathbf{q},\mathbf{T},\delta),
    \end{aligned}
\end{equation}
where $\widetilde{J}_\star$ are various terms of cost function or penalties, and $\lambda_\star$ are relative weights.
Subscripts $\star=\{f,e,t,o,r,d\}$ ($f$) swarm formation similarity, ($e$) denote control effort, ($t$) total time, ($o$) obstacle avoidance, ($r$) swarm reciprocal avoidance, ($d$) dynamic feasibility.
$\delta$ is the sampling time interval.

In our previous work \cite{quan2022formation}, we used the fixed number sampling points $\hat{\mathbf{p}}_{i,j}=p_i((j/\kappa_i)\cdot T_i)$ to transform the optimization problem, where $p_i(t)$ is the $i^{th}$ piece trajectory and $\kappa_i$ is the fixed sample number on this piece.
However, considering that the total time $T_\Sigma$ changes during the optimization process, the fixed number sampling points $\hat{\mathbf{p}}_{i,j}$ are difficult to space on the whole trajectory equally.
Therefore, we take fixed time interval sampling points for the whole trajectory to ensure the accuracy of the penalty function sampling transformation
\begin{equation}
\label{equ:sampling points}
    \begin{aligned}
        & \tilde{\mathbf{p}}_{j}(t)=p_i(j\delta-\sum^{i-1}_{l=1}T_l), \\
        & j\in\{0,\cdots,\kappa\}, \kappa=\lfloor \frac{T_\Sigma}{\delta} \rfloor,
    \end{aligned}
\end{equation}
where $\kappa$ is the sample number and $T_l$ is the preceding time for any $1 \leq l<i$.

For the trajectory planning of swarm robots, the fixed time interval sampling points $\tilde{\mathbf{p}}_{j}(t)$ can simplify the optimization problem.
Compared with $\hat{\mathbf{p}}_{i,j}$, the timestamp corresponding to $\tilde{\mathbf{p}}_{j}(t)$ is fixed, so the states of other robots at this timestamp are also constant during the optimization process.
Therefore, it is feasible to calculate the states of other robots w.r.t $\tilde{\mathbf{p}}_{j}(t)$ according to the broadcast trajectories before optimization.
Then we can solve the uniform formation position sequence optimization (\ref{equ:formation position problem}) in advance and use $\hat{\mathbf{p}}_i^*$ to replace the formation similarity metric $f_s$ in trajectory optimization (\ref{equ_local:cost}) of $i^{th}$ robot.
This decoupled formation trajectory optimization results in higher computational efficiency, making our method suitable for large-scale swarm robots.

Despite the optimization problem is not differentiable when sampling number $\kappa$ changes, the cost function remains continuous w.r.t. time duration $\mathbf{T}$.
In this paper, we use the quasi-Newton method proposed in \cite{lewis2013nonsmooth} to solve the non-smooth discrete unconstrained optimization problem (\ref{equ:unconstainted problem}).

\vspace{-0.3cm}
\subsection{Cost Functions and Gradients}
Given the fixed sampling time interval $\delta$, we can evaluate the cost functions and gradients of the whole trajectory by a finite sum of sampling points $\tilde{\mathbf{p}}_{j}(t)$.
The cost of various general purpose penalties at $j^{th}$ sampling points is 
\begin{equation}
\label{equ:cost}
    \mathcal{P}_\star(\mathbf{c},\mathbf{T},j\delta) = \mathcal{P}_\star(\tilde{\mathbf{p}}_{j}(t)),
\end{equation}
then the cost function $\widetilde{J}_\star$ in (\ref{equ:unconstainted problem}) is calculated as follows
\begin{align}
    \nonumber \widetilde{J}_\star(\mathbf{q}, \mathbf{T},\delta) & = J_\star(\mathbf{c},\mathbf{T},\delta), \\
    &\label{equ:Jx} = \delta \sum_{j=0}^\kappa \bar{\omega}_j \mathcal{P}_\star(\mathbf{c},\mathbf{T},j\delta) + \\
    &\nonumber \frac{1}{2} (T_\Sigma -\kappa\delta)\left[\mathcal{P}_\star(\mathbf{c},\mathbf{T},\kappa\delta) + \mathcal{P}_\star(\mathbf{c},\mathbf{T},T_\Sigma)\right],
\end{align}
where $(\bar{\omega}_0,\bar{\omega}_1,\cdots,\bar{\omega}_{\kappa-1},\bar{\omega}_{\kappa})=(1/2,1,\cdots,1,1/2)$ are the orthogonal coefficients following the trapezoidal rule~\cite{Press2007numerical}.
And MINCO allows any second-order continuous cost function $\widetilde{J}_\star(\mathbf{q},\mathbf{T})$ to be represented by $J_\star(\mathbf{c},\mathbf{T})$.
Hence, $\partial\widetilde{J}_\star/\partial\mathbf{q}$ and $\partial\widetilde{J}_\star/\partial\mathbf{T}$ can be efficiently obtained from $\partial{J}_\star/\partial\mathbf{c}$ and $\partial{J}_\star/\partial\mathbf{T}$ respectively, which is benefit to the construction and solution of the optimization problem.
In (\ref{equ:sampling points}), the sampling time $t=j\delta-\sum^{i-1}_{l=1}T_l$ is related to the preceding time $T_l$, so the gradient of $J_\star$ w.r.t $\mathbf{c}_i$ and $T_l$ are computed as
\begin{equation}
\label{equ:jx_pc}
    \frac{\partial{J}_\star}{\partial \mathbf{c}_i} = \frac{\partial{J}_\star}{\partial \mathcal{P}_\star}                           \frac{\partial \mathcal{P}_\star}{\partial \tilde{\mathbf{p}}_{j}(t)} \frac{\partial \tilde{\mathbf{p}}_{j}(t)}{\partial \mathbf{c}_i},
\end{equation}
\begin{equation}
\label{equ:jx_pT}
    \frac{\partial{J}_\star}{\partial T_l} = \frac{\partial{J}_\star}{\partial \mathcal{P}_\star}                           \frac{\partial \mathcal{P}_\star}{\partial \tilde{\mathbf{p}}_{j}(t)} \frac{\partial \tilde{\mathbf{p}}_{j}(t)}{\partial t} \frac{\partial t}{\partial T_l},
\end{equation}
\begin{equation}
    \frac{\partial \tilde{\mathbf{p}}_{j}(t)}{\partial \mathbf{c}_i} = \boldsymbol{\beta}(t), \frac{\partial \tilde{\mathbf{p}}_{j}(t)}{\partial t}=\dot{\tilde{\mathbf{p}}}_{j}(t), \frac{\partial t}{\partial T_l} = 
	\begin{cases}
		0,  &l=i, \\
		-1, &l<i,
	\end{cases}
\end{equation}
where the calculation of $\partial{J}_\star/\partial \mathcal{P}_\star$ is simple and the details of $\mathcal{P}_\star(\tilde{\mathbf{p}}_{j}(t))$ for various general purpose are given as follow.

\subsubsection{Cost of Swarm Formation Similarity $\mathcal{P}_f$}
In Sec.\ref{subsec:optimization formation}, we decouple the formation similarity error metric from trajectory optimization by constructing an unconstrained optimization problem to calculate the uniform optimal formation position sequence $\hat{\mathbf{p}}_i^*$ for each sampling point.
This improvement avoids multiple calculations of formation similarity metric $f_s$.
Then, we use the quadratic form to calculate the cost of swarm formation similarity
\begin{equation}
\label{equ:P_s}
    \mathcal{P}_f(\tilde{\mathbf{p}}_{j}(t))=\max \{ \parallel \tilde{\mathbf{p}}_{j}(t) - \hat{\mathbf{p}}_{i,j}^* \parallel^2,0 \}^3.
\end{equation}

\subsubsection{Control Effort $J_e$}
\label{subsec:Je}
The $s^{th}$ ($s=3$ in this paper) control input for the trajectory and its gradients are written as 
\begin{equation}
	\label{equ:Je}
	J_e=\sum_{i=1}^{M}\int_{0}^{T_i}\parallel p_i^{(s)}(t)\parallel^2dt,
\end{equation}
\begin{equation}
	\label{equ:Je_c}
	\frac{\partial J_e}{\partial \mathbf{c}_i}=2\left( \int_0^{T_i}\boldsymbol{\beta}^{(s)}(t)\boldsymbol{\beta}^{(s)}(t)^Tdt \right)\mathbf{c}_i,
\end{equation}
\begin{equation}
	\label{equ:Je_T}
	\frac{\partial J_e}{\partial T_i}=\mathbf{c}_i^T\boldsymbol{\beta}^{(s)}(T_i)\boldsymbol{\beta}^{(s)}(T_i)^T\mathbf{c}_i.
\end{equation}

\subsubsection{Total Time $J_t$}
\label{subsec:Jt}
In order to ensure the aggressiveness of the trajectory, we minimize the total time $J_t=\sum_{i=1}^MT_i$. The gradients are given by $\partial J_t/\partial \mathbf{c}=\mathbf{0}$ and $\partial J_t/\partial \mathbf{T}=\mathbf{1}$.

\subsubsection{Cost of Obstacle Avoidance $\mathcal{P}_o$}
\label{subsec:Jo}
Inspired by\cite{zhou2019robust}, obstacle avoidance penalty $J_o$ is computed using Euclidean Signed Distance Field (ESDF). 
We penalize the sampling points which are too close to the obstacles
\begin{equation}
	\label{equ:P_o}
	\mathcal{P}_o(\tilde{\mathbf{p}}_{j}(t)) = \max\{\psi_o(\tilde{\mathbf{p}}_{j}(t)), 0\}^3,
\end{equation}
\begin{equation}
\label{equ:psi_o}
    \psi_o(\tilde{\mathbf{p}}_{j}(t)) = d_o - d_o(\tilde{\mathbf{p}}_{j}(t)), 
\end{equation}
where $d_o$ is the safety threshold set according to the actual situation and $d_o(\tilde{\mathbf{p}}_{j}(t))$ is the distance between $\tilde{\mathbf{p}}_{j}(t)$ and the closest obstacle around it. The gradient of $\mathcal{P}_o$ w.r.t $\tilde{\mathbf{p}}_{j}(t)$ is
\begin{equation}
\label{equ:Po_dp}
    \frac{\partial \mathcal{P}_o}{\partial \tilde{\mathbf{p}}_{j}(t)} = - \nabla d^T,
\end{equation}
where the $\nabla d$ is the gradient of ESDF in $\tilde{\mathbf{p}}_{j}(t)$.

\subsubsection{Cost of Swarm Reciprocal Avoidance $\mathcal{P}_r$}
We penalize $\tilde{\mathbf{p}}_{j}(t)$ when it is too close to the trajectories $p_\phi(t),\phi\in\Phi$ at the fixed timestamp $t=j\delta$, where $\Phi$ represents the all other robots in the swarm.
Compared to our previous work\cite{quan2022formation}, the state of other robots with fixed timestamp $p_\phi(j\delta)$ are constant during the optimization process and do not produce a gradient w.r.t $\mathbf{T}$ for the cost function $J_r$.
So the optimization problem and the gradients are simplified.

The cost of swarm reciprocal avoidance is defined as
\begin{equation}
\label{equ:P_r}
    \mathcal{P}_r(\tilde{\mathbf{p}}_{j}(t))= \sum_{\Phi} \max \{\psi_r(\tilde{\mathbf{p}}_{j}(t),p_\phi(j\delta)),0\}^3,
\end{equation}
\begin{equation}
	\label{eq:phi_d}
	\psi_r(\tilde{\mathbf{p}}_{j}(t),p_\phi(j\delta))=d_r^2-\parallel \tilde{\mathbf{p}}_{j}(t)-p_\phi(j\delta) \parallel^2,
\end{equation}
where $d_r$ is the safe clearance between each robot. 
And the gradient of $\mathcal{P}_r$ w.r.t $\tilde{\mathbf{p}}_{j}(t)$ is
\begin{equation}
\label{equ:Pr_dp}
    \frac{\partial \mathcal{P}_r}{\partial \tilde{\mathbf{p}}_{j}(t)} = -2(\tilde{\mathbf{p}}_{j}(t)-p_\phi(j\delta))^T.
\end{equation}

\subsubsection{Cost of Dynamic feasibility $\mathcal{P}_d$}
We limit the maximum value of velocity and acceleration to guarantee that the robots can execute the trajectory.
\begin{equation}
\label{equ:P_d}
    \begin{aligned}
        \mathcal{P}_d(\tilde{\mathbf{p}}_{j}(t))&=\mathcal{P}_{d,v}(\tilde{\mathbf{p}}_{j}(t))+\mathcal{P}_{d,a}(\tilde{\mathbf{p}}_{j}(t)), \\
        \mathcal{P}_{d,v}(\tilde{\mathbf{p}}_{j}(t)) &= \max \{ \parallel \dot{\tilde{\mathbf{p}}}_{j}(t)\parallel^2 - v_m^2, 0\}^3, \\
        \mathcal{P}_{d,a}(\tilde{\mathbf{p}}_{j}(t)) &= \max \{ \parallel \ddot{\tilde{\mathbf{p}}}_{j}(t)\parallel^2 - a_m^2, 0\}^3,
    \end{aligned}
\end{equation}
where $v_m$ and $a_m$ are the maximum velocity and acceleration.

\subsection{Discussion on solution quality of trajectory optimization}
The proposed trajectory optimization process (\ref{equ:local trajectory problem}) aims to solve a challenging multi-stage Linear Quadratic Minimum Time (LQMT) problem, which is inherently non-convex and non-linear. 
Additionally, incorporating ESDF for obstacle avoidance introduces further non-convex constraints. As a result, guaranteeing the global optimal solution with the quasi-Newton method is not always possible.
To address concerns regarding local minima and infeasible solutions, we have implemented measures that prioritize safety and dynamic feasibility while maintaining high-performance formation flight.

Firstly, we utilize hybrid-A* searching algorithm\cite{zhou2019robust} to generate initial trajectories that are collision-free and dynamically feasible, ensuring a valid final solution trajectory.
During optimization, we give greater weight to obstacle avoidance and dynamic constraints to prioritize safety and feasibility. 
Additionally, we conduct collision checks on trajectories to enhance safety.
Moreover, our distributed swarm optimization framework effectively mitigates the impact of local minima on overall formation performance.
Implementing these measures, our method reliably achieves robust formation flight while maintaining computational efficiency.

\section{Swarm reorganization method}
\label{sec:swarm consensus}
During the formation navigation, the swarm could encounter many unfavorable conditions, such as highly constrained space, inappropriate assignment of tasks, and sudden formation switching commands. To recover from these situations, we present a swarm reorganization method. The method aims to generate high-quality local goals which satisfy the desired formation distribution and respect the current states of each robot. With these local goals, the swarm can reform the desired shape quickly, even in highly constrained environments such as narrow corridors or holes.

Unlike high-frequency distributed formation trajectory optimization, the swarm reorganization method only runs at a low frequency when the stable state of formation flight is destroyed or about to be destroyed.
The method first calculates the formation constraint awareness and then solves an optimal formation \textbf{AL}ignment and task \textbf{AS}signment problem (ALAS).
The former awareness distributedly quantifies the conflict between formation maintenance and obstacle avoidance of each robot, while the latter solves the ALAS problem centrally.

\vspace{-0.5cm}
\subsection{Formation Constraint Awareness}
\label{subsec:awareness}
First, we need to derive weights to indicate how severely the environment constrains the robots. 
These weights are called formation constraint awareness, which should be determined by the current formation status and obstacle information.

Inspired by our previous work~\cite{quan2021eva}, we hope to describe the conflict degree based on the relationship between different gradient information.
Firstly, we retrieve the ESDF distance $d_o(p)$ and the corresponding obstacle gradient $\nabla d(p)$. Meanwhile, we calculate the current gradient $\nabla f_s(p)$ of the formation similarity term. Secondly, we calculate the cosine $\beta$ of the angle between gradients $\nabla d(p)$ and $\nabla f_s(p)$
\begin{equation}
    \beta = \frac{\nabla d(p) \;\nabla f_s(p)}{\parallel\nabla d(p)\parallel\parallel\nabla f_s(p)\parallel}.
\end{equation}
Then we utilize the sigmoid function to map the cosine of the angle to a conflict coefficient $\eta$
\begin{equation}
\label{alas_equ}
    \eta(\beta) = \frac{1}{1+e^{(\alpha\beta+\gamma)}},
\end{equation}
where $\alpha$ regulates how fast this awareness rises as the cosine value $\beta$ increases, $\gamma$ regulates the dead zone and the activation zone of this angle-based awareness.
The conflict coefficient $\eta$ is maximum when the directions of $\nabla d(p)$ and $\nabla f_s(p)$ are opposite, which indicates the most conflicting case. And $\eta$ reaches a minimum when the two gradients have the same direction, which means no conflict.

The formation constraint awareness should also consider the influence of the gradient magnitude and the distance of the current closest obstacle. Hence, we design the formation constraint awareness $g_i$ of $i^{th}$ robot as
\begin{equation}
\label{equ:constrained aware}
    g_i = \lambda\cdot \eta(\beta)\cdot\frac{\parallel\triangledown J_f(p)\parallel}{d_o(p)}.
\end{equation}
We apply the calculation to each robot in the swarm and thus get a formation constraint awareness vector $\mathbf{g}=\{g_1,\dots,g_N\}$ of the whole swarm. 
To distinguish the most constrained ones,  we use $\textrm{softmax}$ function to amplify the variance of awareness vector $\mathbf{g}$ and normalize the vector
\begin{equation}
    \mathbf{w} = \textrm{softmax}(\mathbf{g}).
\end{equation}
$\lambda$ in (\ref{equ:constrained aware}) is used to adjust the variance of elements in $\mathbf{w}$. 
$\mathbf{w}$ is the final awareness vector describing the degree of formation-obstacle conflict of the robots in the swarm.

\begin{figure}
    \begin{center}
        \includegraphics[width=1.0\columnwidth]{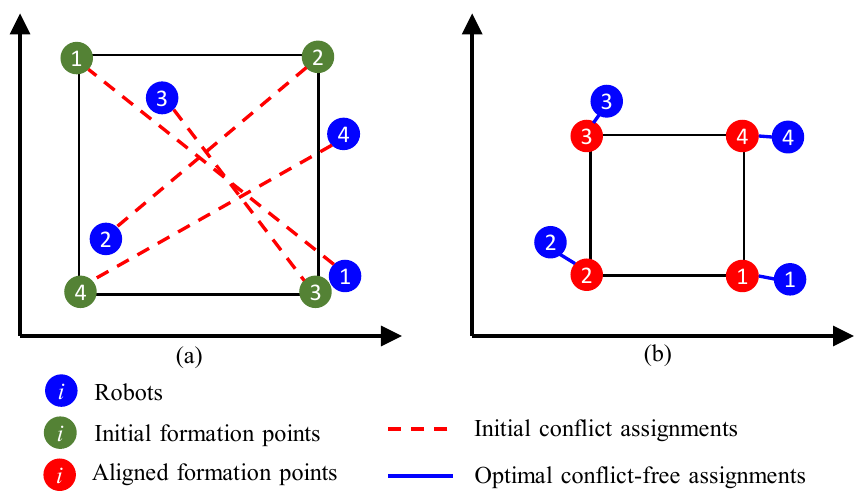}
    \end{center}
    \vspace{-0.5cm}
    \caption{Illustration of formation alignment and task assignment. (a) Before solving ALAS problem, the initial formation goals suffer from a large transition distance to robots and disordered assignments that may lead to deadlock. (b) With robots at the same positions, after solving ALAS, the formation goals enjoy low distance costs and better assignments.}
    \label{fig:alas}
    \vspace{-0.5cm}
\end{figure}

\begin{figure}
    \begin{center}
        \includegraphics[width=1.0\columnwidth]{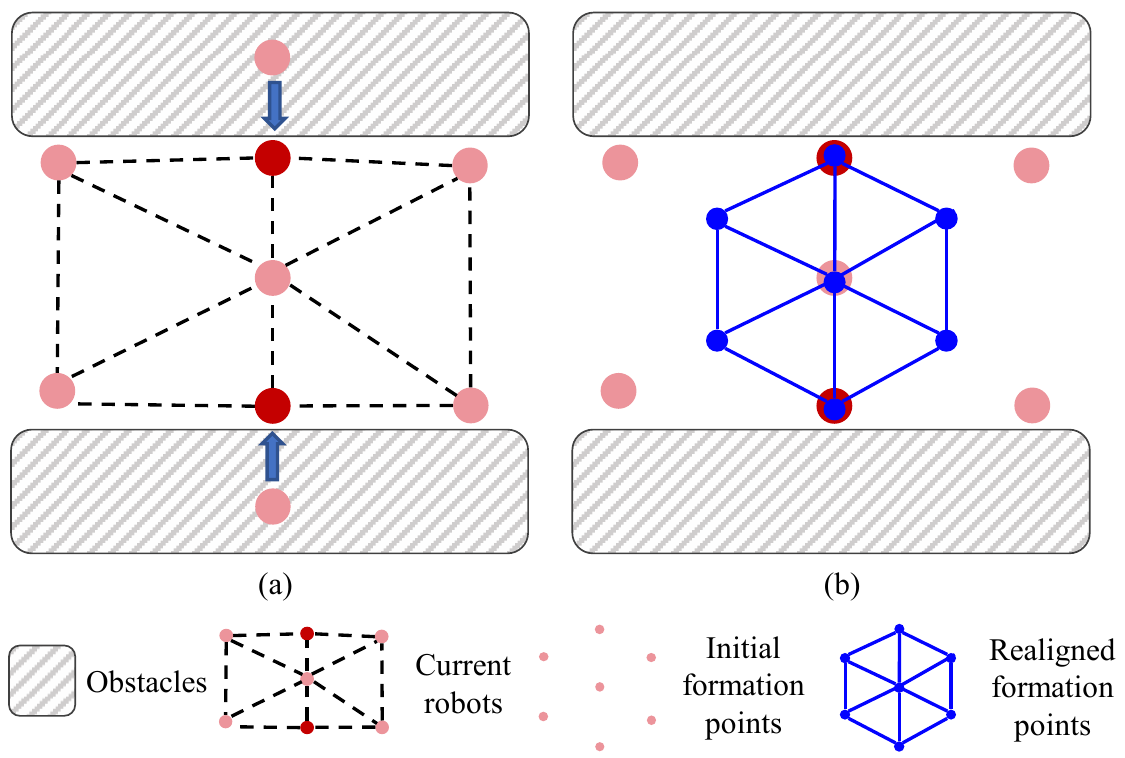}
    \end{center}
    \vspace{-0.5cm}
    \caption{Illustration of the weighted formation alignment problem in a narrow corridor. When a hexagon formation enters the corridor, the original formation distribution(pink points) cannot be maintained anymore. The upper and lower robots(red points) are severely pressed by the obstacles, and hence have the largest constraint awareness. After solving the awareness-weighted formation alignment, the swarm obtains a new desired formation distribution(blue points) that best matches up with the constrained robots.}
    \label{fig:alignment}
    \vspace{-0.5cm}
\end{figure}

\vspace{-0.5cm}
\subsection{Formation Alignment and Task Assignment Optimization}
For formation flights in constrained scenes, e.g. in narrow corridors, the unconstrained robots with lower constraint awareness possess larger space to freely coordinate with other robots, since the obstacles don't hinder the formation requirement. On the contrary, the constrained robots with larger awareness always fall into the conflict between formation maintenance and obstacle avoidance. 
Hence, refining the positions of unconstrained robots to match up with the constrained ones is more reasonable when generating local goals for formation reorganization. 
In this work, we use $\mathbf{w}$ in Sec.\ref{subsec:awareness} to weigh the robots when adjusting the formation distributions and place more weights on the constrained robots.

In this section, we only elaborate on the ALAS problem for local goal generation. Afterward, the global-remap-local-replan strategy uses the generated local goals to reorganize the formation, which is detailed in Sec.\ref{subsec:GRLR-strategy}. 

Let $\mathbf{lg}_i = [lg_{ix},lg_{iy},lg_{iz}]^T,i = 1,...,N$ represents the current positions of robots. 
The desired formation shape template is given by $N$ positions $\mathbf{q}_j = [q_{jx},q_{jy},q_{jz}]^T,j = 1,...,N$. 
Then, the aligned formation positions $\mathbf{q}'_j$ can be written as
\begin{equation}
    \mathbf{q}'_j = s\cdot\mathbf{q}_j + \mathbf{d},
\end{equation}
where $s\in\mathbb{R}$ is the scaling factor, $\mathbf{d}\in\mathbb{R}^3$ denotes the translation factor.
In this work, the formation alignment is determined by a scaling factor and a translation factor.

ALAS is composed of formation alignment and task assignment as shown in Fig.~\ref{fig:alas}.
The former aims to find the optimal alignment of the desired formation based on a weighted Euclidean distance cost. 
And the latter solves the optimal assignment that matches the agents with the local goals.

The task assignment problem is formulated as 
\begin{equation}
\label{equ:assignment problem}
    \min_{\sigma}\sum_i^n \parallel \mathbf{lg}_i - (s^*\cdot\mathbf{q}_{\sigma(i)} + \mathbf{d}^*)\parallel^2,
\end{equation}
where $\sigma\in S_N$ is the assignment map of the formation task and $S_N$ is the symmetric group of all permutations from the set $\{1,...,N\}$ to itself. Problem (\ref{equ:assignment problem}) solves the assignment that minimizes the overall transition distance between current robots and the aligned local goals. 

The formation alignment problem is formulated as
\begin{equation}
\label{equ:alignment problem}
    \min_{s,\mathbf{d}}\sum_i^n w_i \cdot \parallel \mathbf{lg}_i - (s\cdot\mathbf{q}_{\sigma^{*}(i)} + \mathbf{d})\parallel^2,
\end{equation}
where $\sigma^{*}$ represents the optimal assignment, $w_i$ is the awareness weight from Sec.\ref{subsec:awareness}.  Problem (\ref{equ:alignment problem}) generates a standard formation that best fits into the current robot positions according to a distance cost weighted by the constraint awareness. Fig.~\ref{fig:alignment} illustrates how the alignment adjusts the formation distribution when the swarm is traversing a corridor.

Problems (\ref{equ:assignment problem}) and (\ref{equ:alignment problem}) are coupled. The whole ALAS problem has three decision variables: scaling factor $s$, translation $\mathbf{d}$, and assignment $\sigma$. The goal of ALAS is to find an optimal set of decision variables that minimize both (\ref{equ:alignment problem}) and (\ref{equ:assignment problem}). Note that there is no awareness weight $w_i$ multiplied in formulation (\ref{equ:assignment problem}). Because in the assignment optimization, we only care about the total Euclidean distance cost, which is irrelevant to the degree of the constraint of any agent. 

However, for formation alignment using only scaling factor $s$ and translation $\mathbf{d}$, \cite{agarwal2018simultaneous} proves that the corresponding assignment $\sigma$ can be optimized in a decoupled manner, rather than alternating the two optimization phases iteratively. In \cite{agarwal2018simultaneous}, the optimal assignment solution $\sigma^*$ is shown invariant w.r.t the changes in formation scaling factor $s$ and translation $\mathbf{d}$.
And the solution of (\ref{equ:assignment problem}) can be directly optimized by solving the following integer programming with new pseudo costs $\kappa_{ij}$
\begin{equation}
\begin{aligned}
\label{equ:task cost}
     &\min_{\sigma=(x_{ij})}\sum_{i=1}^n\sum_{j=1}^n \kappa_{ij}x_{ij},\\
  \;&\text{where}\;\;\kappa_{ij} = -\mathbf{lg}_i^T \mathbf{q}_j.
\end{aligned}
\end{equation}

The formulation (\ref{equ:task cost}) is independent of the scale parameter $s$ and translation $\mathbf{d}$. Hence, (\ref{equ:task cost}) can be first solved prior to the formation alignment phase. Then we determine the best alignment using the optimized assignment $\sigma^*$. 
The formation alignment problem with awareness weights is still convex and the closed-form solution to (\ref{equ:assignment problem}) is given In Appendix.\ref{app:alignment}.

Given the solution of ALAS, the position of generated local goal $\mathbf{lg}'_i$ for the $i^{th}$ robot is calculated by
\begin{equation}
\label{equ:gen local}
    \mathbf{lg}'_i = s^{*}\cdot\mathbf{q}_{\sigma^{*}(i)} + \mathbf{d}^{*}.
\end{equation}
After the ALAS optimization, the distribution of generated local goals is in the desired formation shape, and respects the formation-obstacle conflict of each robot.

\section{Swarm agreement method}
\label{sec:swarm agree}

\subsection{Global-remap-local-replan Strategy}
\label{subsec:GRLR-strategy}
\begin{figure}
    \begin{center}
        \includegraphics[width=0.9\columnwidth]{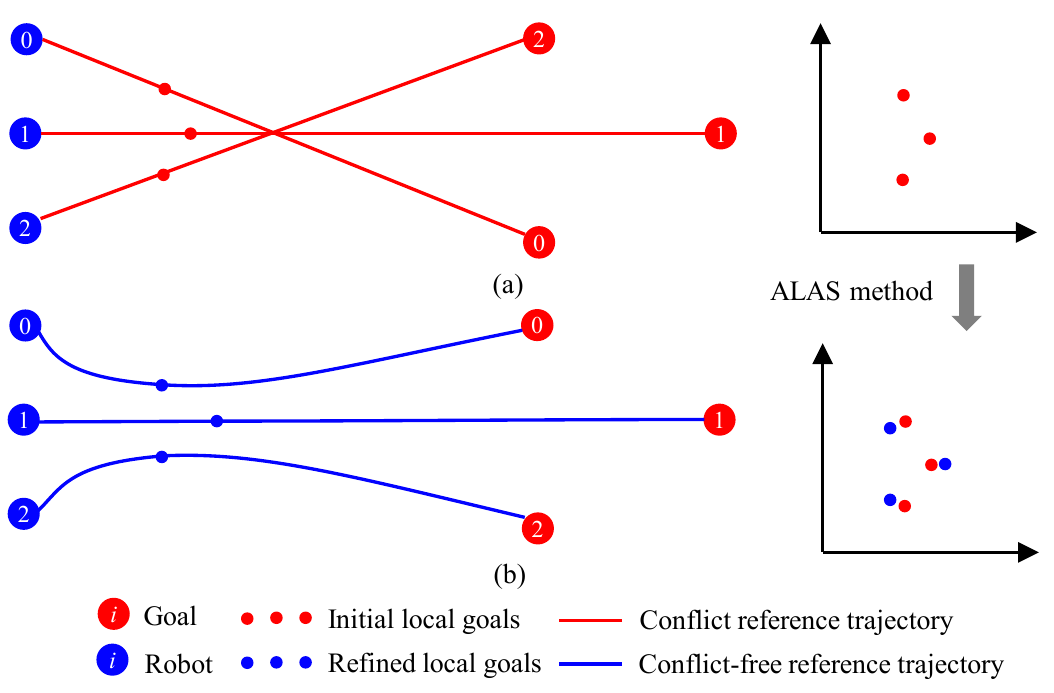}
    \end{center}
    \vspace{-0.5cm}
    \caption{Illustration of GRLR strategy. (a) The local-replan strategy generates initial local goals within the planning horizon. Due to the conflict reference trajectories, the swarm robots are expected to deadlock with the current formation behavior. Therefore, the global-remap strategy calls ALAS method to realign the shape and reassign the task of initial local goals. (b) After solving the refined local goals, the global-remap strategy generates conflict-free reference trajectories. In this way, the swarm formation will converge quickly with the new formation behavior.}
    \label{fig:grlr}
\end{figure}

\begin{algorithm}
\label{grlr}
    \begin{algorithmic}[1]
        \Require Global reference trajectory $\mathcal{T}_{ref}$, \, Local trajectory $\mathcal{P}_i$, \, Initial local goals $\mathbf{lg}_i$, \, Global goals $\mathbf{G}_i$, \, Assignments $\sigma$, \, Scale $s$, \, Translation $\mathbf{d}$ ;\\
        
        \textbf{Initialize:} $CallGlobalRemap \leftarrow False$,
        \For{each robot i}
        \State $g_i \leftarrow$ \textbf{ConstrainedAwareness(}$\mathcal{P}_i$\textbf{)}; \Comment{detailed in (\ref{equ:constrained aware})}
        \If{$g_i>g_d$} \Comment{$g_d$ is threshold of $g_i$}
        \State $CallGlobalRemap \leftarrow True$;
        \State $\mathbf{w}\leftarrow \textrm{softmax}(\mathbf{g})$;
        \EndIf
        \EndFor
        \If{\textbf{SimilarityError(}$\cdot$\textbf{)}$>e_{sim,d}$} \Comment{detailed in (\ref{equ:L_metric})}
        \State $CallGlobalRemap \leftarrow True$;
        \State $\mathbf{w}\leftarrow \textrm{softmax}(\mathbf{I})$;
        \EndIf
        \If{$CallGlobalRemap$}
        \State $\sigma^*\leftarrow$ \textbf{Assignment($\sigma, s, \mathbf{d}$)};\Comment{detailed in (\ref{equ:assignment problem})}
        \State $s^*,\mathbf{d}^*\leftarrow$\textbf{Alignment($\mathbf{w},\sigma^*$)};\Comment{detailed in (\ref{equ:alignment problem})}
        \State $\mathbf{lg}_i^*\leftarrow$\textbf{RemapLocalGoals(}$\sigma^*,s^*,\mathbf{d}^*$\textbf{)};
        \State $\mathbf{G}_i^*\leftarrow$\textbf{RemapGlobalGoals(}$\mathbf{G}_i,\sigma^*$\textbf{)};
        \State $\mathcal{T}_{ref}^*\leftarrow$\textbf{GlobalTrajectoryReplan(}$\mathbf{lg}_i^*,\mathbf{G}_i^*$\textbf{)};
        \State \textbf{Return} $\mathcal{T}_{ref}^*$;
        \EndIf
    \end{algorithmic}
\caption{Global-remap strategy}
\end{algorithm}

The swarm system cannot quickly recover from disordered states caused by unknown obstacles or sudden changes in the desired formation shape.
To address this challenge, we propose a novel approach that utilizes a global-remap-local-remap (GRLR) strategy for trajectory generation. 
This approach enables us to efficiently navigate complex environments while maintaining the swarm in a ``critical state" of the desired formation, effectively balancing coordination and adaptability.

For distributed framework, communication helps the robot to obtain information about others and generate better coordination behavior.
Especially in the case of formation transformation, collaborative decision-making can make the swarm formation converge quickly.
However, circular dependencies sometimes occur due to communication delays, making it difficult to guarantee the consistency of decisions.
Therefore, We design a \textbf{G}lobal-\textbf{R}emap-\textbf{L}ocal-\textbf{R}eplan (GRLR) strategy for the swarm formation system, which only centrally remaps crucial parameters of formation and maintains the formation coordination through distributed replanning local trajectory.

GRLR strategy comprises the local-replan for a single robot and the global-remap for a formation-level system.
The local-replan is a receding horizon incremental planning strategy\cite{fei2017iros}, which allows each robot plans a trajectory within its limited sensing range.
The local goals are selected on the global reference trajectories within planning horizon $\Psi_p$, as shown in Fig.~\ref{fig:grlr} (a).
The global-remap is an efficient centralized strategy that only remaps the local goals by solving ALAS method and refines the global reference trajectory, as shown in Fig.~\ref{fig:grlr} (b).
GRLR strategy is very suitable for distributed asynchronous systems, and there is no deadlock in swarm systems even in the presence of network delays.

The main workflow of the proposed global remap strategy is described in Algorithm.\ref{grlr}.
Before generating the new formation behavior, the global-remap strategy checks if there are any emergence events (Line 1-12), such as the stable state of the swarm formation being destroyed (Line 9) or about to be destroyed (Line 4).
Unlike the local-replan strategy is triggered at a fixed frequency, the global-remap strategy is started by emergent events (Line 13).
Then the ALAS method is called to solve the optimal assignment $\sigma^*$ and alignment $s^*, \mathbf{d}^*$ (Line 14-15).
Global-remap strategy remaps the local goals $\mathbf{lg}_i^*$ and global goals $\mathbf{G}_i^*$  and generates a new global trajectory $\mathcal{T}_{ref}^*$ for each robot (Line 16-19).
Finally, robots form the new formation by executing the local-replan strategy.

In this work, we utilize this semi-distributed GRLR strategy to make the swarm formation adaptable to unknown obstacles or sudden changes in the desired formation shape by replanning local trajectories at 1 Hz and checking emergent events for triggering the global-remap strategy at 20 Hz.

\vspace{-0.5cm}
\subsection{Formation-level Global Path Finding}
\label{subsec:global-path}
\begin{algorithm}
\caption{Formation-level Global Path Finding}
\label{alg1}
\begin{algorithmic}[1]
	\Require Tree\ $\mathcal{T}$,\; State\ $\mathbf{z}$,\; Path cost $c$,\; Path $\mathbf{P}$;\\
	
	\textbf{Initialize:} $\mathcal{T}_a \leftarrow \emptyset \cup \{\textbf{z}_{start}\}$,\, $\mathcal{T}_b \leftarrow \emptyset \cup \{\textbf{z}_{goal}\}$,\;\qquad$c_{best}\leftarrow \infty$,\, $FoundSolution \leftarrow False$;
			
	\For{$i = 1$ to $N$}
	\State $\textbf{z}_{random} \leftarrow$ \textbf{Sample(}$\textbf{z}_{start},\,\textbf{z}_{goal},\,c_{best}$\textbf{)};
	\If{\textbf{not} $FoundSolution$ }
    \State $\textbf{z}_{new} \leftarrow$  \textbf{GreedyExtendTree(}$\mathcal{T}_a,\,\textbf{z}_{random}$\textbf{)};
    \State $\textbf{z}_{conn} \leftarrow$ \textbf{NearestVertice(}$\textbf{z}_{new},\,\mathcal{T}_b$\textbf{)};
    \State $c_{new} \leftarrow$ 
    \textbf{Connect(}$\textbf{z}_{new},\,\textbf{z}_{conn},\,\mathcal{T}_a,\,\mathcal{T}_b$\textbf{)};
    \If{$c_{new} < c_{best}$}
    \State $c_{best} \leftarrow c_{new}$; 
    \State $FoundSolution \leftarrow True$;
    \EndIf
    \Else
    \State $\textbf{z}_{new} \leftarrow$ 
    \textbf{ExtendTree(}$\mathcal{T}_a,\,\textbf{z}_{random}$\textbf{)};
    \State $\mathcal{Z}_{near} \leftarrow$ \textbf{NearVertex(}$\textbf{z}_{new},\,\mathcal{T}_{a}$\textbf{)};
    \State $\mathcal{T}_{a} \leftarrow$ \textbf{Rewire(}$\textbf{z}_{new}$,$\mathcal{Z}_{near}$\textbf{)};
    \State $\textbf{z}_{conn} \leftarrow$  
    \textbf{NearestVertice(}$\textbf{z}_{new},\,\mathcal{T}_b$\textbf{)};
    \State $c_{new} \leftarrow$ \textbf{Connect(}$\textbf{z}_{new},\,\textbf{z}_{conn},\,\mathcal{T}_a,\,\mathcal{T}_b$\textbf{)};
    \If{$c_{new} < c_{best}$} 
    \State $c_{best} \leftarrow c_{new}$;
    \EndIf
	\EndIf
	\State \textbf{SwapTrees(}$\mathcal{T}_a,\,\mathcal{T}_b$\textbf{)};
	\EndFor
	\\$\mathbf{P} \leftarrow $  \textbf{RetrievePath(}$\mathcal{T}_a,\,\mathcal{T}_b$\textbf{)};
	\\\textbf{Return} $\mathbf{P}$;
	\end{algorithmic}
\end{algorithm} 

We propose a method for formation-level global path finding. 
Given a start and goal configuration, the planner generates a feasible path connecting them with collision-free intermediate formations. 
A bidirectional RRT approach is employed to address this path-finding problem. 

Many navigation tasks expect the formation to maneuver with a desired scale. In practice, an oversized formation could reduce the vehicle's communication quality, while an overly small formation scale could increase the risk of inter-vehicle collisions. 
Unlike the method in \cite{baaberg2017formation} which only samples position $\mathbf{p} \in \mathbb{R}^3$ of the formation center, our method adds the formation scale $s$ into the sampling space and makes the formation configuration $\mathbf{z} = \{\mathbf{p},s\} \in \mathbb{R}^3 \times \mathbb{R}^+ $. In this way, the objective of maintaining desired scale, i.e., minimizing the changes in scale along the path, can be handled by minimizing the $L_{2}$-norm distance of path in the configuration space $\mathbf{z}$.

Navigation in dense environments requires the robots to maintain a formation while letting the obstacles pass through the formation. 
The method in \cite{alonso2017multi} samples the center position $\mathbf{p}$, and then the scale factor $s$ is solved by optimizing the formation placement in obstacle-free convex regions. 
However, this approach does not allow any obstacle to intersect with the convex hull of the formation and hence wastes many solutions. 
In contrast, our method directly samples the whole states of the formation configuration $\mathbf{z}$ to fully explore the solution space. 
For each edge of our RRT algorithm, a collision check is conducted on each robot rather than the formation's whole convex hull to allow obstacles to pass. 

The main workflow of our bidirectional RRT planner is described in Algorithm.\ref{alg1}, where two trees $\mathcal{T}_a$ and $\mathcal{T}_b$ grow towards each other from the initial state $\textbf{z}_{start}$ and the goal state $\textbf{z}_{goal}$ respectively. 
Before the first solution is found, the bidirectional planner extends the trees in an RRT-Connect\cite{kuffner2000rrt} manner (Line 5-7). In \textbf{GreedyExtendTree()} and \textbf{Connect()}, the greedy heuristic\cite{kuffner2000rrt} is adopted to aggressively explore the environment and make tree-connection attempts. 
After a feasible solution is found, i.e. a finite path cost $c_{new}$ is returned by \textbf{Connect()}, the function \textbf{Sample()} computes an informed sampling set with the new cost $c_{new}$ as depicted in \cite{gammell2014informed}. 
Then the standard Bidirectional-RRT*\cite{qureshi2015intelligent} procedures are conducted in each loop to update the trees (Line 13-17). 
Since the path cost is $L_{2}$-norm distance in the configuration space $\mathbf{z}$, informed sampling\cite{gammell2014informed} and Bidirectional-RRT*\cite{qureshi2015intelligent} can guarantee the asymptotic optimality of the path solution. 

This formation-level path planner is deployed to render a global path when the global environment information is available. Then global trajectories connecting the waypoints of the global path are generated using MINCO\cite{wang2021geometrically}, and the framework in Sec.\ref{sec:trajectory optimization} is employed to optimize the local motions. 

\section{Benchmark}
\label{sec:benchmark}
In the benchmark, it is important to assess the distortion degree of the current formation $\mathcal{F}^c$ fairly relative to the desired one $\mathcal{F}^d$ during flight.
Inspired by \cite{parker2018pipeline}, we solve the following nonlinear optimization problem to find the best similarity transformation ($Sim(3)$ transformation) that aligns $\mathcal{F}^c$ with $\mathcal{F}^d$.
Then the average formation distance degree $\overline{e}_{dist}$ is calculated at the normalized formation scale
\begin{equation}
    \label{equ:e_dist} 
    \overline{e}_{dist} = \frac{1}{s_o \cdot L}\int_{\mathcal{L}} \displaystyle{\min_{\mathbf{R},\,\mathbf{t},\,s} \sum_{i=1}^n ||\mathbf{p}_{i}^d -  (s\,\mathbf{R}\,\mathbf{p}^\mathbf{c}_i + \mathbf{t})||^2}dl,
\end{equation}
where $\mathbf{p}_{i}^d$ and $\mathbf{p}_{i}^c$ represent the position of $i^{th}$ robot in formation $\mathcal{F}^d$ and $\mathcal{F}^c$, respectively.
The $Sim(3)$ transformation is composed of a rotation $\mathbf{R} \in SO(3)$, a translation $\mathbf{t}\in \mathbb{R}^3$ and a scale expansion $s \in \mathbb{R}_+$.
Moreover, $s_o$ is the initial formation scale, and $L$ is the length of formation trajectory $\mathcal{L}$.
Optimizing the transformation in (\ref{equ:e_dist}) and applying it to formations, the influence of scaling and rotation is squeezed out so that all the formations can be equitably rated by measuring the position error w.r.t the desired formation. 
A larger $\overline{e}_{dist}$ represents a larger distortion from the desired formation $\mathcal{F}^d$.
Besides, we also calculate the average formation similarity degree $\overline{e}_{sim}$
\begin{equation}
    \label{equ:e_sim}
    \overline{e}_{sim} = \frac{1}{s_o \cdot L}\int_{\mathcal{L}} \parallel\mathbf{\hat{L}}-\mathbf{\hat{L}}_{des}\parallel^2_Fdl,
\end{equation}
where $\mathbf{\hat{L}}$ and $\mathbf{\hat{L}}_{des}$ are detailed in (\ref{equ:L_metric}) and the formation similarity error $\parallel\mathbf{\hat{L}}-\mathbf{\hat{L}}_{des}\parallel^2_F$ is proposed in Sec.\ref{sec:formation}.
We show important parameters in Table~\ref{tab:parameter} used in the following benchmarks, simulations, and real-world experiments.
All benchmarks are run on a desktop with an Intel i7-12700 CPU.

\begin{table}
    \centering
    \caption{Formation parameters of the proposed method}
    \label{tab:parameter}
    \renewcommand\arraystretch{1.2}
    \begin{tabular}{l c c}
        \toprule
        Parameter    & Symbol  &Value      \\
\hline
    Similarity error threshold  & $e_{sim,d}$               & 0.05      \\
    Constraint awareness threshold  & $g_d$               & $2/N$       \\
    Parameter for regulation in (\ref{alas_equ})    & $\alpha$             & 5     \\
    Parameter for regulation in (\ref{alas_equ})    & $\lambda$  & 25      \\ 
    Parameter for regulation in (\ref{alas_equ})    & $\gamma$   &  -1     \\
    Sampling time interval ($s$)      & $\delta$   & 0.5    \\
    Planing Horizon ($m$)         & $\Psi_p$  & 7.5 \\
    Max velocity ($m/s$)          & $v_m$  & 1.0 \\
    Max acceleration ($m/s^2$)    & $a_m$  & 6.0 \\
    Weight for control effort & $\lambda_e$  & 10000.0 \\
    Weight for total time & $\lambda_t$                 & 80.0 \\
    Weight for swarm reciprocal avoidance & $\lambda_r$      & 10000.0 \\
    Weight for obstacle avoidance & $\lambda_o$ & 10000.0 \\
    Weight for swarm formation similarity & $\lambda_f$ & 10000.0 \\
    Weight for dynamic feasibility & $\lambda_d$ & 10000.0 \\
       \toprule
    \end{tabular}
\end{table}

\begin{figure}
    \begin{center}
        \includegraphics[width=0.9\columnwidth]{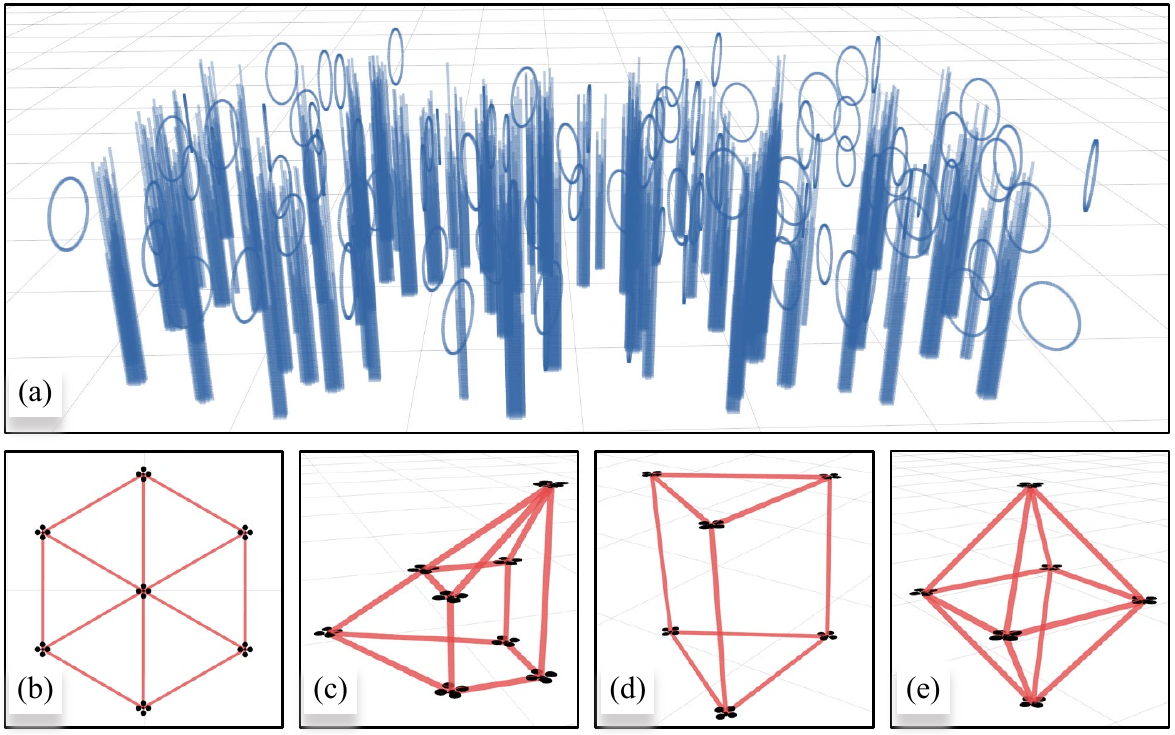}
    \end{center}
    \vspace{-0.5cm}
    \caption{Random forest map and formation types of benchmarks. (a) Random forest map. (b) Regular hexagon shape. (c) Irregular shape. (d) Triangular prism shape. (e) Octahedron shape.}
    \label{fig:benchmark_map}
    \vspace{-0.6cm}
\end{figure}

\begin{table*}
    \caption{Performance Comparison between Formation Definition Methods}
        \label{tab:adaptability_benchmark}
        \centering
        \begin{tabularx}{\textwidth}{@{}l*{15}{>{\centering\arraybackslash}X}c@{}}
        \toprule
            & \multicolumn{1}{c}{\textbf{Formation type}} & \multicolumn{2}{c}{Regular hexagon} & \multicolumn{2}{c}{Irregular shape} & \multicolumn{2}{c}{Triangular prism}  & \multicolumn{2}{c}{Octahedron}   \\
        \toprule
        {\textbf{Scenario}}  & \diagbox[width=8em]{Method}{Error} & $\overline{e}_{dist}(\%)$ & $\overline{e}_{sim}(\%)$ & $\overline{e}_{dist}(\%)$ & $\overline{e}_{sim}(\%)$ & $\overline{e}_{dist}(\%)$ & $\overline{e}_{sim}(\%)$ & $\overline{e}_{dist}(\%)$ & $\overline{e}_{sim}$(\%) \\
        \toprule
        \multirow{4}{*}{\textbf{Same to same}}
        & Position & 39.120 & 0.384& 35.649 & 0.384 & 34.506 & 0.374  & 21.534 & 0.341  \\
        & Displacement & 16.231 & 0.172 & 15.023 & 0.159 & 14.952 & 0.125 & 11.645 & 0.153  \\
        & Distance  & 15.489 & 0.164& 14.295 & 0.131   & \textbf{14.009} & \textbf{0.118}& 10.285 & 0.113   \\
        & \textbf{Ours} & \textbf{15.443} & \textbf{0.161} & \textbf{14.287} & \textbf{0.128}& 14.012 & 0.119 & \textbf{10.281} & \textbf{0.112}  \\
        \midrule
        \multirow{4}{*}{\textbf{Rotation change}}
        & Position & 57.456 & 0.945 & 51.298 & 0.732& 49.821 & 0.612 & 34.124 & 0.542  \\
        & Displacement  & 39.456 & 0.412& 31.012 & 0.439& 29.546 & 0.345& 23.125 & 0.353   \\
        & Distance  & 27.513 & 0.312& 22.312 & 0.234& 21.031 & 0.201  & 14.173 & 0.159 \\
        & \textbf{Ours}& \textbf{19.234}  & \textbf{0.218} & \textbf{17.032} & \textbf{0.171}& \textbf{15.013} & \textbf{0.151} & \textbf{12.146} & \textbf{0.138}   \\
        \midrule
        \multirow{4}{*}{\textbf{Scale change}} 
        & Position  & 59.654 & 1.098& 58.416 & 0.784& 53.246 & 0.741 & 37.845 & 0.555  \\
        & Displacement  & 42.516 & 0.629& 40.021 & 0.624 & 39.412 & 0.398 & 29.845 & 0.395 \\
        & Distance & 59.542 & 1.030& 59.105 & 0.799& 54.126 & 0.632 & 38.451 & 0.578   \\
        & \textbf{Ours} & \textbf{18.332}  & \textbf{0.192} & \textbf{18.196} & \textbf{0.185}& \textbf{16.023} & \textbf{0.179}& \textbf{12.264} & \textbf{0.164}   \\
        \midrule
        \multirow{4}{*}{\textbf{Scale \& rotation change}} 
        & Position  & 62.584 & 1.304 & 60.124 & 0.796& 56.213 & 0.832& 41.856 & 0.635  \\
        & Displacement & 45.627 & 0.755 & 40.194 & 0.631& 40.168 & 0.423  & 31.288 & 0.504 \\
        & Distance  & 62.154 & 1.250& 61.059 & 0.804& 54.317 & 0.684& 42.138 & 0.684   \\
        & \textbf{Ours}  & \textbf{20.231}  & \textbf{0.243}& \textbf{18.345} & \textbf{0.204}& \textbf{16.851} & \textbf{0.183}& \textbf{12.357} & \textbf{0.175}   \\
       \bottomrule
       \end{tabularx} 
\end{table*}

\begin{table*}
    \caption{Performance Comparison between Formation Navigation Methods}
    \label{tab:predictability_benchmark}
    \centering
    \begin{tabularx}{\textwidth}{@{}l*{15}{>{\centering\arraybackslash}X}c@{}}
    \toprule
        & \multicolumn{1}{c}{\textbf{Formation type}} & \multicolumn{4}{c}{Regular hexagon}   \\
    \toprule
    {\textbf{Scenario}}  & \diagbox[width=8em]{Method}{Error} & $success \quad rate(\%)$ & $length(m)$ & $\overline{e}_{dist}(\%)$ & $\overline{e}_{sim}(\%)$ \\
    \toprule
    \multirow{4}{*}{\textbf{Sparse}}
    & VRB\cite{zhou2018agile} & 75 & 22.978& 57.962 & 0.984   \\
    & Spatial-only & 100 &21.923 & 15.023 & 0.152   \\
    & \textbf{Spatial-temporal} & \textbf{100} & \textbf{21.756} & \textbf{11.240} & \textbf{0.138}  \\
        \midrule
    
    \multirow{4}{*}{\textbf{Medium}}
    & VRB\cite{zhou2018agile} &25  & - & - & - \\
    & Spatial-only  & 100 & 22.130& 14.927 & 0.158  \\
    & \textbf{Spatial-temporal}& \textbf{100}  & \textbf{21.932} & \textbf{13.274} & \textbf{0.153}  \\
    \midrule
    
    \multirow{4}{*}{\textbf{Dense}} 
    & VRB\cite{zhou2018agile}  & 0 & -& - & - \\
    & Spatial-only  & 100 & 22.283& 17.630 & 0.185 \\
    & \textbf{Spatial-temporal}  & \textbf{100}  & \textbf{22.133}& \textbf{15.443} & \textbf{0.161} \\
   \bottomrule
    \end{tabularx} 
\end{table*}

\vspace{-0.5cm}
\subsection{Adaptability of Graph-based Formation Definition}
\label{subsec:adaptability_benchmark}
To demonstrate the adaptability of graph-based formation definition in Sec.\ref{sec:formation}, we conduct numerous benchmarks compared to the mainstream formation definition methods concluded in\cite{oh2015survey}, which are categorized based on the controlled variables, namely position-based\cite{turpin2012trajectory}, distance-based\cite{olfati2004consensus} and displacement-based methods\cite{krick2009stabilisation}.

We implement these methods in our framework and adapt them to the dense environments by replacing the original cost $J_s$ in (\ref{equ:formation position problem}) to generate uniform optimal formation position sequence $\hat{\mathbf{p}}_i^*$ for each robot $i$.
For the position-based method, we set drone\_0 as the leader and predefined the absolute relative positions for all other robots to specify the desired formation.
So its cost is $J_{s,1}=0$.
The distance-based method optimizes the error of desired inter-agent distances
\begin{equation}
    \label{equ:J_s_2}
        J_{s,2} = \sum_{j \in N  } \left(\left\|\mathbf{p}_{i}-\mathbf{p}_{j}\right\|-\left\|\mathbf{p}_{i}^{d}-\mathbf{p}_{j}^{d}\right\|\right)^{2},
\end{equation}
where $N$ is the number of robots, and $\mathbf{p}_{i}^{d}$ is the desired position vector for the $i^{th}$ robot.
The displacement-based method optimizes the error of desired relative displacements
\begin{equation}
    \label{equ:J_s_3}
        J_{s,3} = \sum_{j \in N  } \left\| (\mathbf{p}_{i}-\mathbf{p}_{j})-(\mathbf{p}_{i}^{d}-\mathbf{p}_{j}^{d})\right\|^2.
\end{equation}

Then we simulate four different geometric formation types in a high-density environment of $40\times15m$ size with randomly generated obstacles, as shown in Fig.~\ref{fig:benchmark_map} (a).
2D and 3D formations with irregular and regular geometries are considered, namely formation types in a regular hexagon, irregular geometry, triangular prism, and octahedron, as shown in Fig.~\ref{fig:benchmark_map} (b)-(e).
To fully compare the adaptability of these methods, we design four different scenarios considering both scaling and rotational variation of formation shape.
The formation's initial and final positions may differ in scale and rotation.
Then the scenarios are corresponding categorized as 'Same to same', 'Rotation change', 'Scale change', and 'Scale \& rotation change'.
We test each method 20 times for each scenario and formation type.
The corresponding results over $\overline{e}_{dist}$ (\ref{equ:e_dist}) and $\overline{e}_{sim}$ (\ref{equ:e_sim}) are summarized in Table~\ref{tab:adaptability_benchmark}. 

Unlike our graph-based formation definition, in other methods, changing the scale and rotation of the formation is not permitted during the flight.
As shown in Table~\ref{tab:adaptability_benchmark}, in the same formation type and same scenario, the data states that our method achieves promising results with almost the lowest $\overline{e}_{sim}$ and $\overline{e}_{dist}$.
Moreover, our method shows the lowest error growth rate when the scenario becomes more complicated.
In the same scenario, the distortion degrees of all methods decrease with the change of formation type from 2D to 3D centrosymmetric structure, which shows that the formation maintenance is also related to the structural stability of the formation itself.
In addition, the distance-based method is invariant to the rotation and achieves relatively acceptable performance in the 'Rotation change' scenario.
Nevertheless, it can not handle size-variant cases.
Similarly, other methods are sensitive to rotation or scaling, leading to significant performance degradation in such scenarios.
Generally speaking, our graph-based formation definition method achieves scaling and rotational invariance.
The invariance improves the formation flight's adaptability and outperforms the mainstream methods in complicated scenarios.

\begin{figure}
    \begin{center}
        \includegraphics[width=0.9\columnwidth]{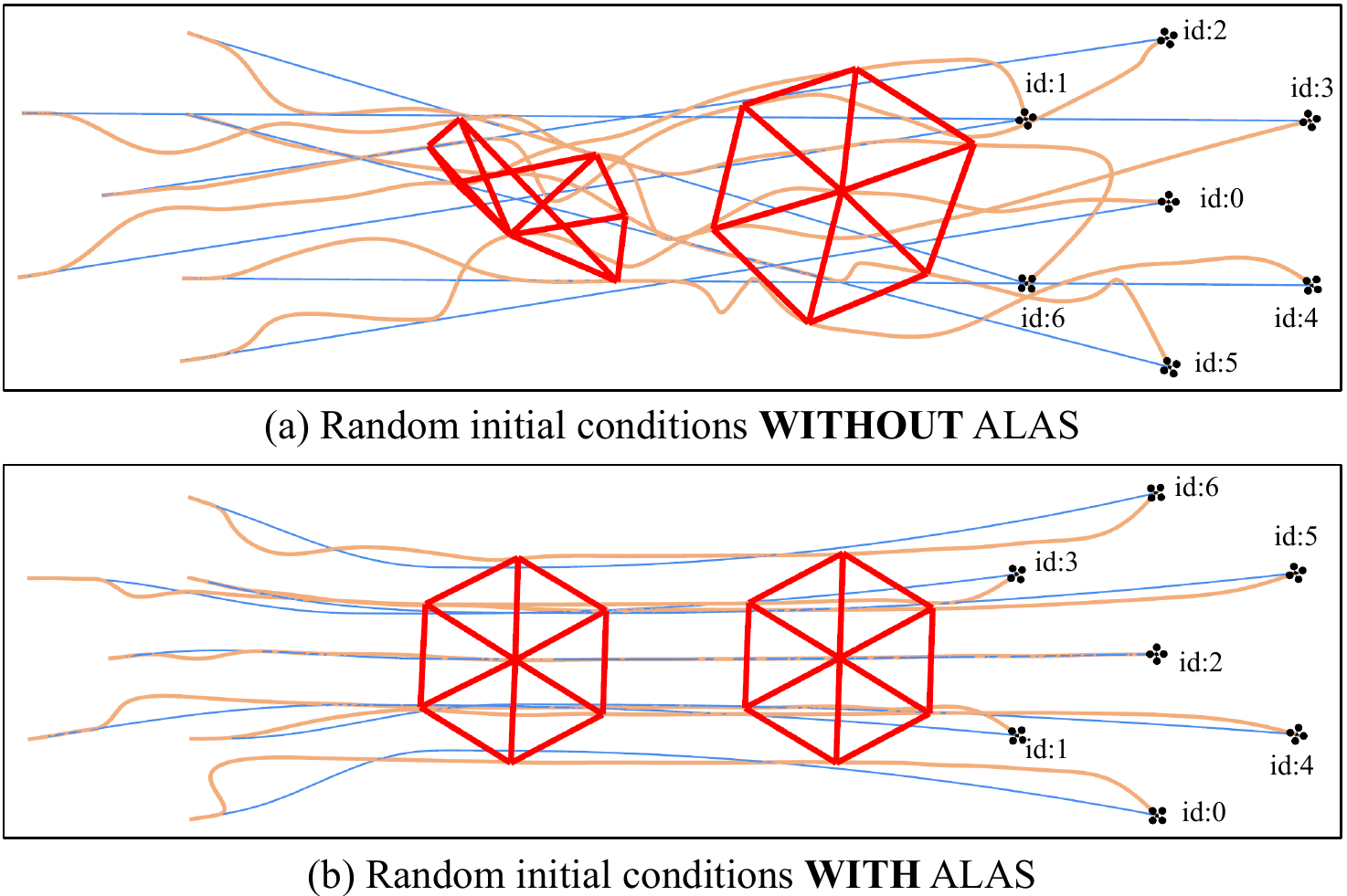}
    \end{center}
    \vspace{-0.5cm}
    \caption{Comparison of formation flight under improper initial conditions with and without ALAS. (a) In the case without ALAS, the executed trajectories (orange lines) are winding, and the formation shape (red lines) converges slowly due to the crossed global trajectories (blue lines). (b) In the case of ALAS, swarm reorganization makes the formation flight process orderly.}
    \label{fig:random_alas}
    \vspace{-0.5cm}
\end{figure}

\begin{figure*}
    \begin{center}
        \includegraphics[width=1.8\columnwidth]{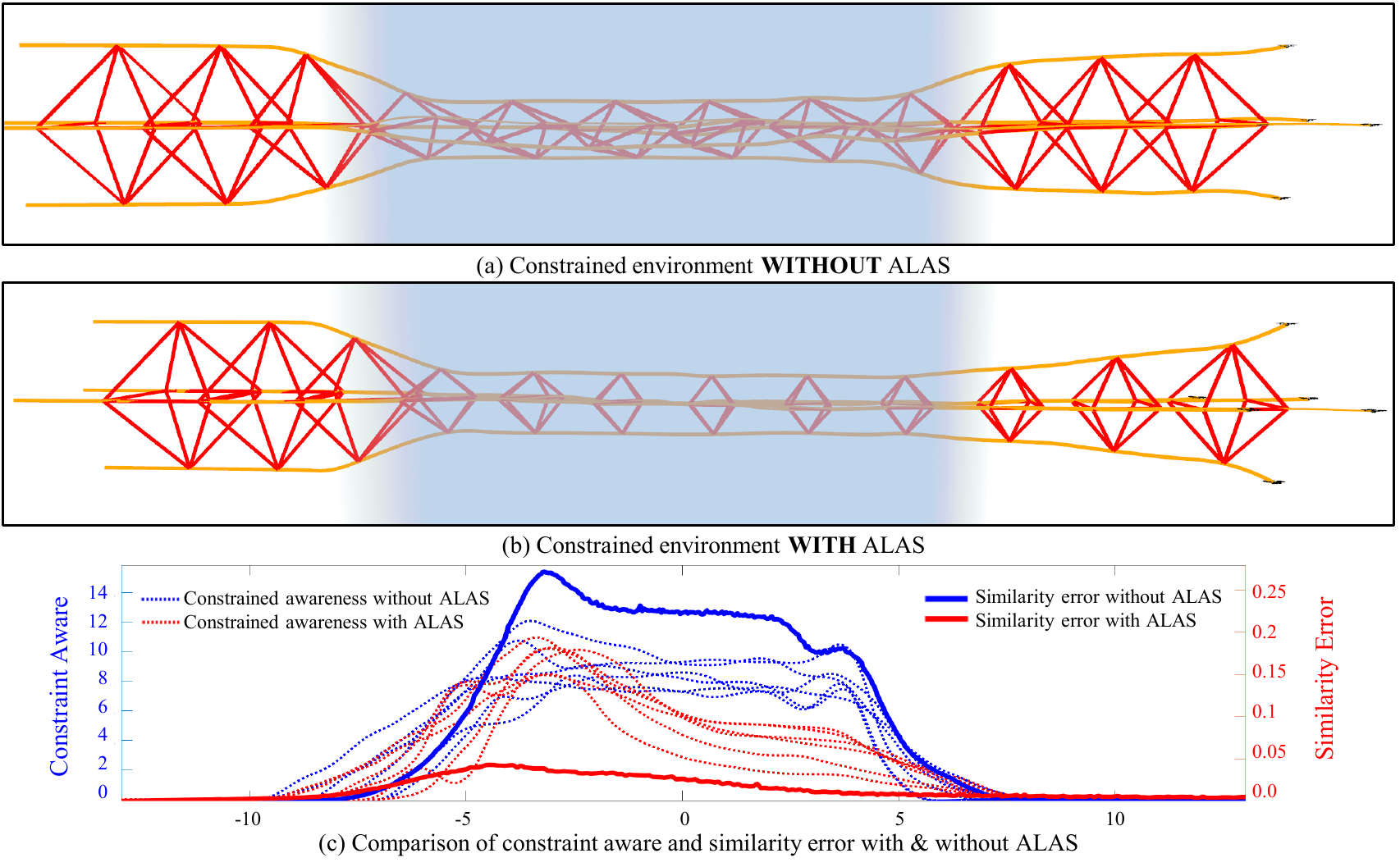}
    \end{center}
    \vspace{-0.5cm}
    \caption{Comparison of formation flight in a constrained environment with and without ALAS. The light blue area represents a wall with a hole in which swarm robots should shrink to pass through. (a) In the case without ALAS, the formation shape (red lines) is deformed to force through the area. (b) In the case with ALAS, the formation shape actively gets smaller to pass the area. (c) Formation flight with ALAS can decrease each robot's constraint awareness (red dotted lines). This improvement maintains the formation shape with lower similarity error (red lines).}
    \vspace{-0.5cm}
    \label{fig:constraint_alas}
\end{figure*}

\vspace{-0.5cm}
\subsection{Predictability of Spatial-Temporal Trajectory Optimization}
To prove the predictability of formation trajectory optimization in Sec.\ref{sec:trajectory optimization}, we compare our work with the virtual rigid body (VRB) method \cite{zhou2018agile}, a SOTA formation control framework that avoids obstacles using potential fields.
Moreover, we also compare the performance between the spatial-only and spatial-temporal optimization to illustrate the importance of the time domain for formation flight.
We simulate seven drones flying in a regular hexagon from one side to another with a velocity limit of $0.5m/s$.
The cluttered area is of $30\times15m$ size, and three obstacle densities are tested for comparison. 
Parameters are finely tuned for the best performance of each method.

The results are summarized in Table~\ref{tab:predictability_benchmark}, which indicates that the VRB method \cite{zhou2018agile} has an unsatisfactory success rate when dealing with medium and dense obstacles.
This is mainly due to the short-term obstacle avoidance generated by multiple interacting potential fields, which often leads to local minima near the corridors, causing robots to become trapped and fail.
However, optimization methods consider the future movement of formation, so they can balance the formation maintenance and obstacle avoidance but not break the formation shape.
Therefore, optimization methods achieve better performance and maintain the success rate.

We can also conclude that the spatial-temporal method is much more effort-efficient, robust, and flexible when considering temporal optimization.
The spatial-only method cannot adjust the trajectory in the time domain, which leads to excessive spatial deformations of the trajectory.
So the trajectory length and the formation error $\overline{e}_{sim}$ and $\overline{e}_{dist}$ are larger in the spatial-only method.

\vspace{-0.5cm}
\subsection{Elasticity of Swarm Reorganization Method}
To validate the swarm reorganization methods in Sec.\ref{sec:swarm consensus}, we design two benchmarks to illustrate the necessity of task assignment and the adaptability of formation alignment.

Firstly we design a comparison of regular hexagon formation flight under improper initial conditions with and without ALAS to validate the necessity of formation task assignment, as shown in Fig.~\ref{fig:random_alas}.
The blue lines represent each robot's global trajectory and its assigned tasks in the formation.
In Fig.~\ref{fig:random_alas}(a), the global trajectories are partially crossed due to inappropriate task assignment, leading to trajectory optimization conflicts.
So the executed trajectories shown by orange lines look very disordered, and the formation shape shown by red lines converges slowly.
In Fig.~\ref{fig:random_alas}(b), the above problems are effectively resolved by considering ALAS.
After one calculation of ALAS, the swarm robots reassign formation tasks and quickly reach a swarm consensus.
Then the swarm formation smoothly converges to the desired shape and navigates to the destination in an energy-efficient way.
The results of this test validate the necessity of task assignment.

Then, we compare formation flight with and without ALAS when passing through a constrained hole to display the adaptability of ALAS.
The results in Fig.~\ref{fig:constraint_alas}(a) and Fig.~\ref{fig:constraint_alas}(b) show that the formation shape may be deformed when passing through the corridor without ALAS.
Otherwise, the case with ALAS can adaptively adjust the formation shape to the constrained environments.
From the quantitative analysis results in Fig.~\ref{fig:constraint_alas}(c), the case with ALAS can quickly adjust the formation scale and make the swarm reach a consensus so that the formation similarity error and constraint awareness of each robot decline rapidly.
However, in the case without ALAS, a higher similarity error and constraint awareness are maintained until the swarm formation leaves the hole, which means the swarm formation is always within the limitations of the environment so that the formation shape cannot converge. 
This benchmark proves the adaptability of formation alignment.

\begin{table}
    \centering
    \caption{Comparison between Global path finding methods}
    \label{tab:group-level path searching}
    \renewcommand\arraystretch{1.2}
    \begin{tabular}{l c c}
        \toprule
                          & Alonso-Mora's method\cite{alonso2017multi} & Ours \\ \hline
        Sampling time $(s)$ & 2.0               & 2.0       \\
        Desired scale $(m)$ & 3.0               & 3.0       \\
        Path length   $(m)$ & 50.77             & \textbf{24.10}     \\
        Min scale along the path $(m)$ & 1.88   & \textbf{2.98}      \\ \toprule
    \end{tabular}
\end{table}

\begin{figure}
    \begin{center}
        \includegraphics[width=0.9\columnwidth]{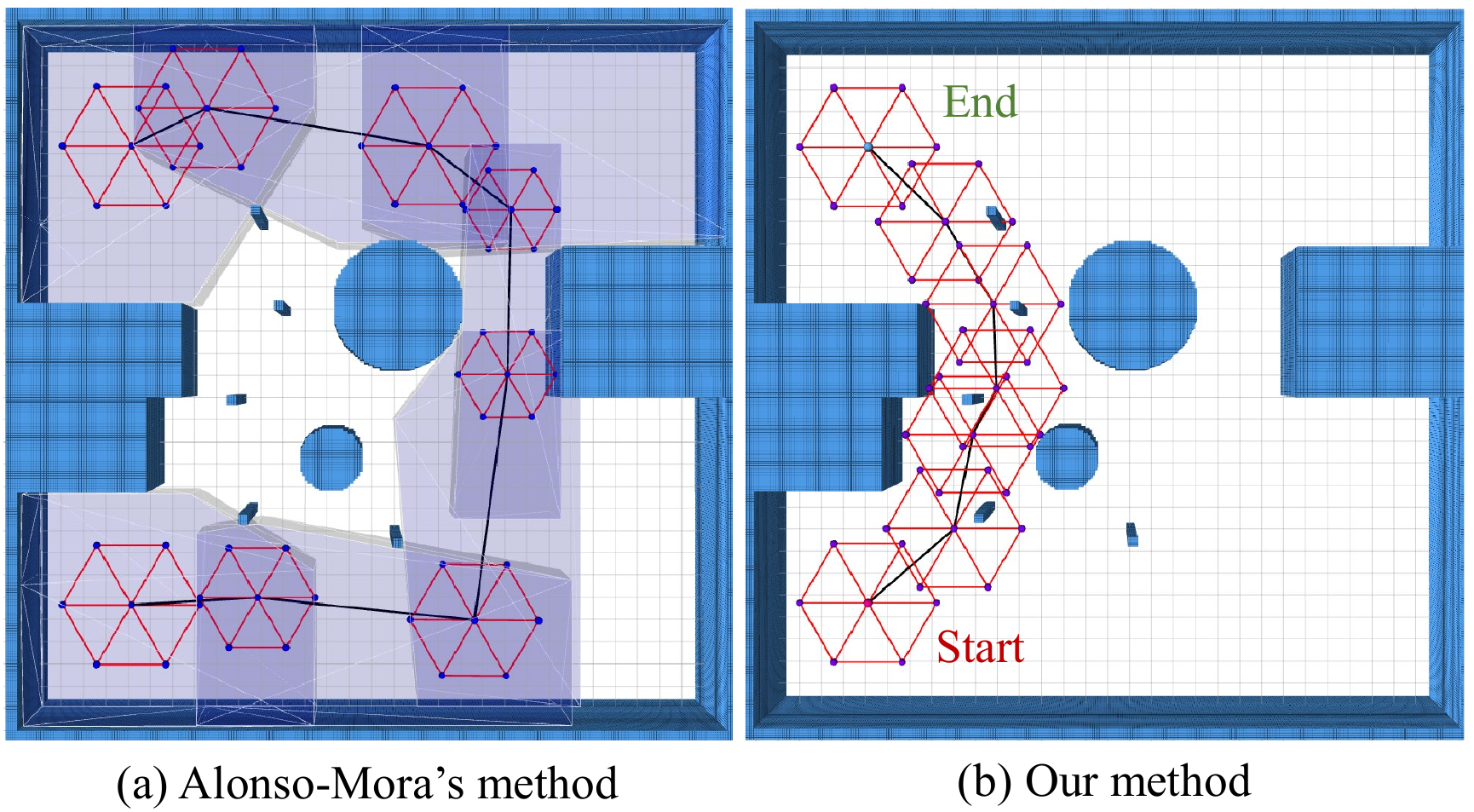}
    \end{center}
    \vspace{-0.5cm}
    \caption{Comparison of Alonso-mora's method and our formation-level path finding method in the same constrained map. (a) This method samples convex regions (purple polyhedra) in free space and connects them if the intersections are traversable in formation. Because the convex regions must be generated in the safe space, this method is too conservative to generate a longer path with a smaller scale. (b) Our method directly samples the whole states of the 3D-scale formation configuration to fully explore the solution space to allow the formation to pass through tiny obstacles.}
    \vspace{-0.5cm}
    \label{fig:global_benchmark}
\end{figure}

\begin{figure}
    \begin{center}
        \includegraphics[width=0.9\columnwidth]{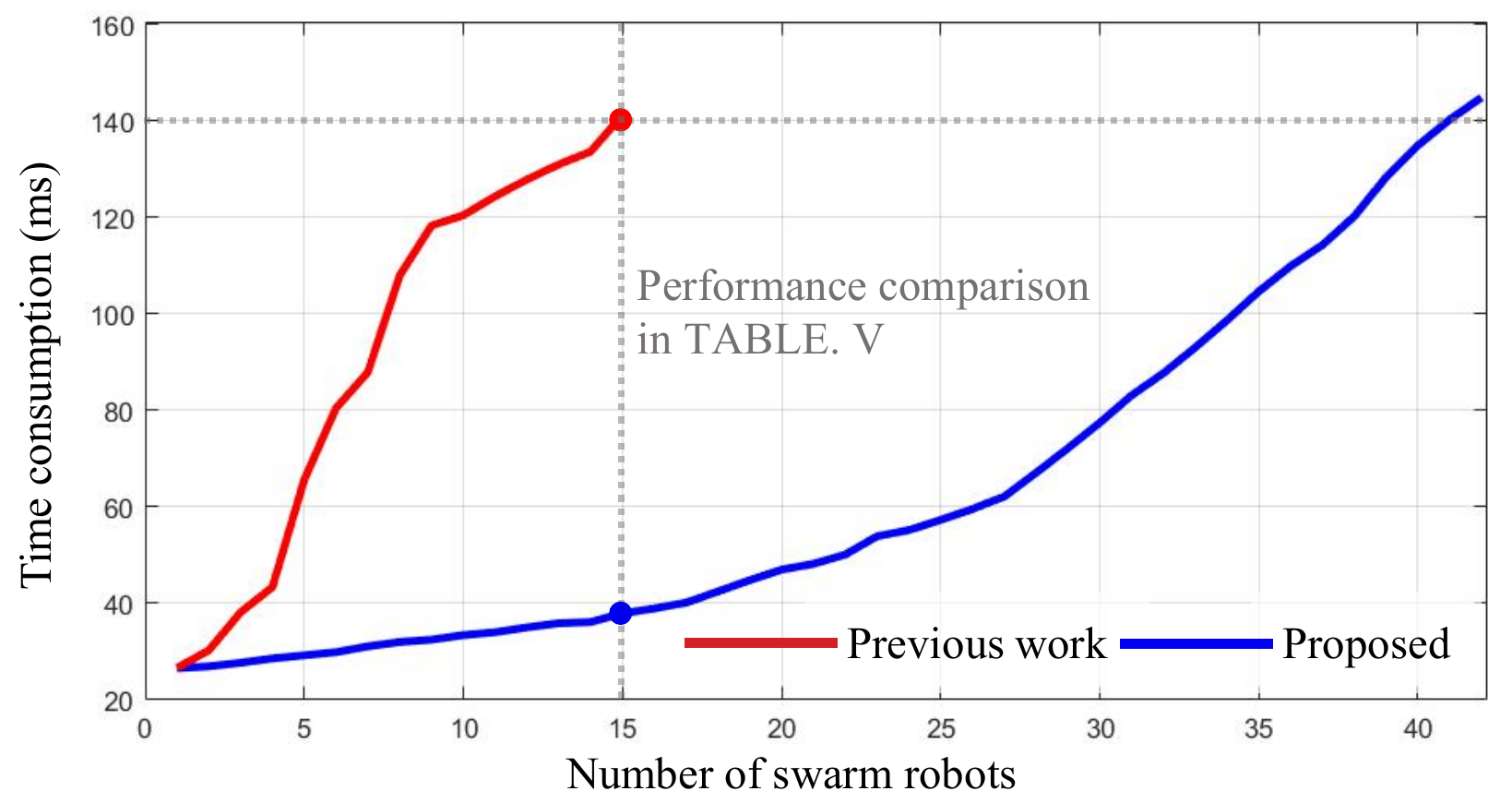}
    \end{center}
    \vspace{-0.5cm}
    \caption{Comparison of time consumption between previous work and proposed work \cite{quan2022formation}. We compare the performance of two methods in the 15-robots scenario. The detailed comparison results are shown in Table~\ref{tab:proposed_method}.}
    \label{fig:time}
\end{figure}

\vspace{-0.5cm}
\subsection{Resilience of Swarm Agreement Method}
To highlight the resilience of our swarm agreement method, we compare Alonso-Mora's global planning method\cite{alonso2017multi} and our formation-level path finding method in a constrained map, as shown in Fig.~\ref{fig:global_benchmark}.
This map comprises several blocks and some tiny obstacles, in which swarm robots need to find the path that allows the formation to pass safely. 
We test the planners 20 times, and Table~\ref{tab:group-level path searching} shows the averaged resultant data.
The results state that Alonso-Mora's method\cite{alonso2017multi} is unsatisfactory in dealing with this scenario. 
Method \cite{alonso2017multi} has no penalty for scale changes of formation, which could choose the corridor route that leads to sudden changes in formation scale.
Moreover, it can not handle tiny obstacles and thus yield to a longer path with smaller scales, as shown in Fig.~\ref{fig:global_benchmark}(a).
Unlike Alonso-Mora's method, our formation-level path-finding method directly samples in the augmented 3D-scale space and can better maintain the desired formation scale.
As shown in Fig.~\ref{fig:global_benchmark}(b), our method generates a much shorter path and only sacrifices a small quantity of formation scale.
Therefore, our method can handle the map with blocks and tiny obstacles and find safe guidance for swarm formation, which is more suitable for dense environments.

\vspace{-0.5cm}
\subsection{Efficiency of Decoupled Formation Optimization}
\begin{table}[t]
    \centering
    \caption{Performance comparison in 15-drones scenario}
    \label{tab:proposed_method}
    \renewcommand\arraystretch{1.2}
    \begin{tabular}{l c c}
        \toprule
                          & Previous method\cite{quan2022formation} & Proposed method\\ \hline
        Time consumption $(ms)$ & 141.7          & \textbf{38.2}  \\
        success rate $(\%)$ & 95.0           & \textbf{100.0}  \\
        length  $(m)$ & 47.387               & \textbf{45.282} \\
        $\overline{e}_{dist}$ $(\%)$ & 12.439 & \textbf{11.724}  \\
        $\overline{e}_{sim}$ $(\%)$ & 0.147   & \textbf{0.139}      \\ \toprule
    \end{tabular}
\end{table}
We compare our proposed decoupled formation optimization with the previously coupled formation optimization\cite{quan2022formation} which directly calls formation similarity distance metric (\ref{equ:L_metric}) multiple times in the optimization process.
To ensure a fair comparison, we exclude the ALAS problem during this benchmark.
Both methods' results are shown in Fig.~\ref{fig:time}.
The previous method\cite{quan2022formation} only supports small-scale swarm formation since the computation time grows exponentially.
Thanks to the decoupled approach, the time consumption of our proposed method for a swarm of 42 robots is not more than 150ms, which can support the real-time application for large-scale swarms.

We select the experimental data from the 15-drones scenario, as presented in Table~\ref{tab:proposed_method}. Our method not only achieves significantly shorter computation times than the previous method, but also demonstrates better performance in terms of formation maintenance, success rate, and trajectory length. 
This validates the effective decoupling of our method, leading to comprehensive performance improvements.

\section{Real world experiments and simulation }
\label{sec:experiments}
\subsection{Real world Experiments}
\begin{figure}
    \begin{center}
         \includegraphics[width=0.9\columnwidth]{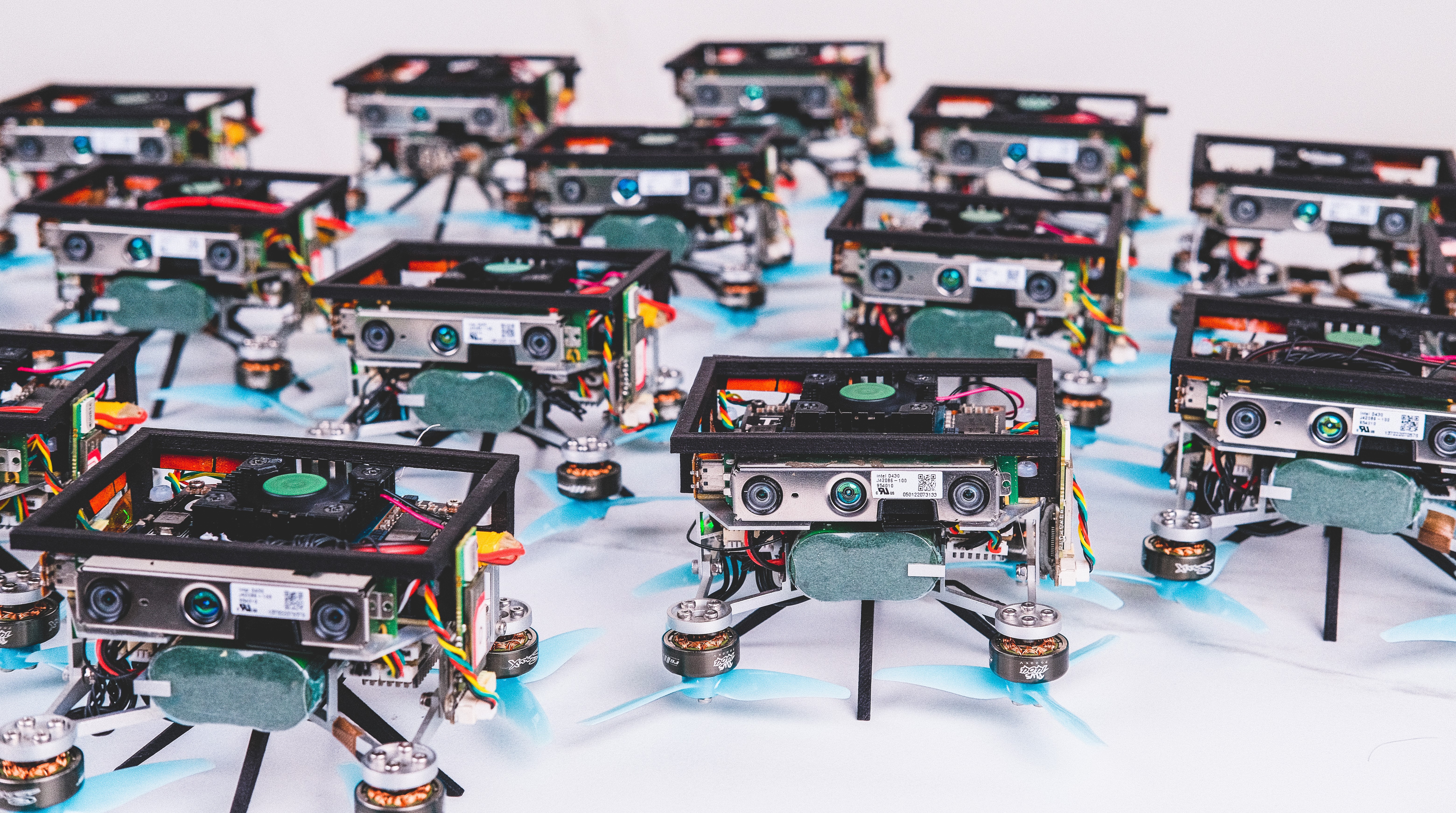}
    \end{center}
    \vspace{-0.5cm}
    \caption{Illustration of palm-sized swarm aerial robots.}\label{fig:real_system}
    \vspace{-0.3cm}
\end{figure}

Our method is integrated with an autonomous distributed aerial swarm system stated in Sec.\ref{sec:system}. 
The swarm shares some information, such as trajectories, through a broadcast network, which is the only connection among all robots.
As shown in Fig.~\ref{fig:real_system}, we use a palm-sized quadrotor platform~\cite{zhou2022swarm} with local sensors and an onboard computer.
Software modules such as estimation, perception, planning, and control are all running onboard in real-time.
The maximum number of swarm robots during real-world experiments is 16.
Three different real-world experiments are designed to verify the proposed formation flight system's characteristics fully.

\begin{figure}
    \begin{center}
         \includegraphics[width=0.9\columnwidth]{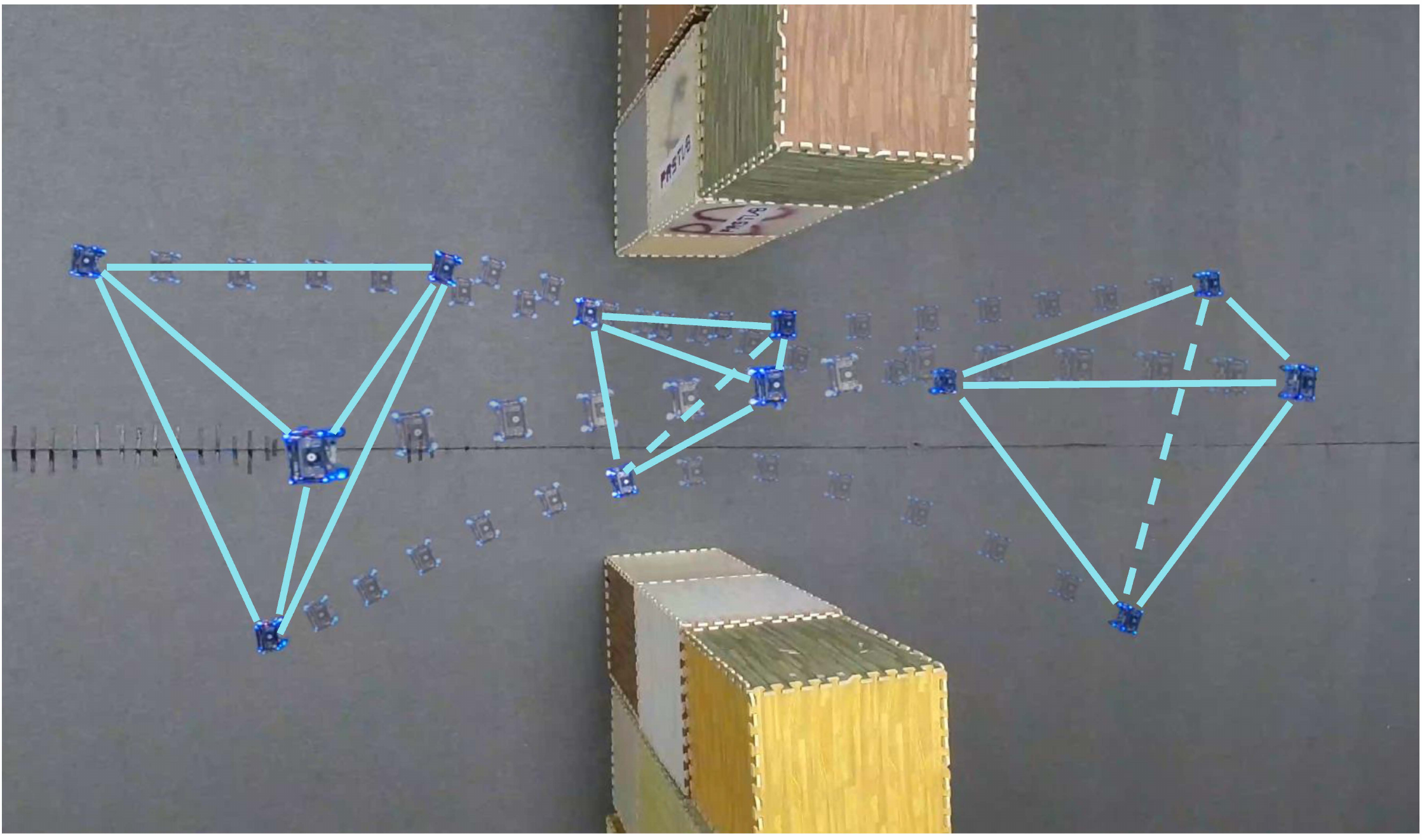}
    \end{center}
    \vspace{-0.5cm}
    \caption{Composite snapshots of a regular tetrahedron formation passing through a corridor. The swarm rotates and compresses the formation shape to fly through the narrow space from right to left. The blue line shows the outline of the formation shape.}\label{fig:real_4}
    \vspace{-0.5cm}
\end{figure}

\begin{figure*}[t]
    \begin{center}
         \includegraphics[width=1.8\columnwidth]{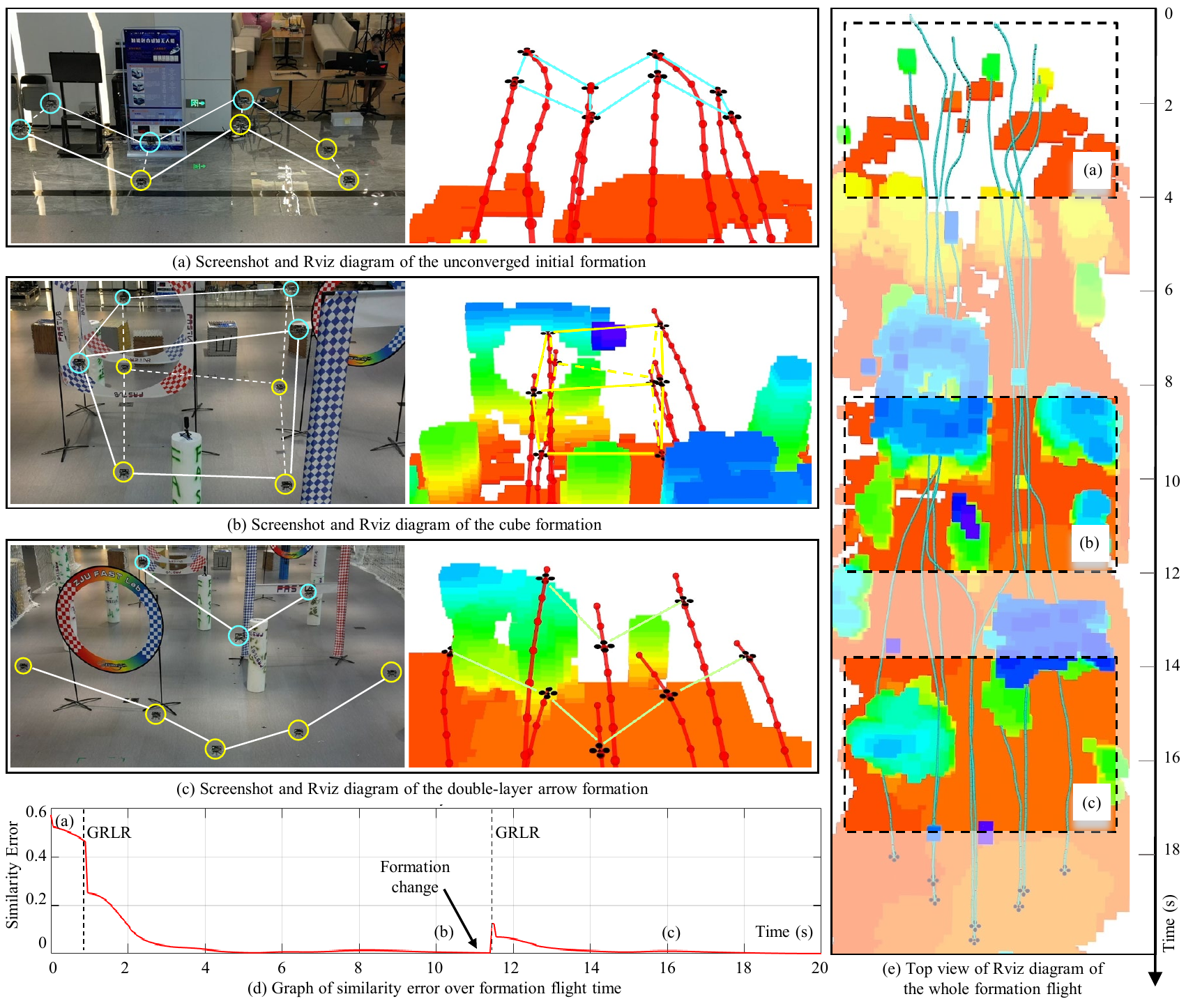}
    \end{center}
    \vspace{-0.5cm}
    \caption{Illustration of 3-D formation transformation experiment. The blue circle represents quadrotors assigned to the upper position, and the yellow circle represents a lower position. The white line represents the outline of the formation shape. (a) Formation flight starts from unconverged initial positions and improper initial task assignments. So swarm robots call the GRLR strategy to reorganize the formation parameters. (b) After 3 seconds, swarm robots converge to the desired cube shape. Then swarm receives a formation transformation command and quickly calls the GRLR strategy. (c) After 3 seconds, swarm robots converge to the desired double-arrow shape. (e) The executed trajectories (light blue lines) indicate that swarm formation is convergent most time. (d) A more accurate numerical analysis states that similarity error decreases quickly after calling the GRLR strategy.}\label{fig:real_8}
    \vspace{-0.5cm}
\end{figure*}

In the first experiment, as shown in Fig.~\ref{fig:real_4}, four quadrotors in a 3-D regular tetrahedron formation manage to pass through a narrow corridor safely.
During the flight, the swarm adaptively rotates and compresses the formation shape in response to environmental changes. This test proves that the scaling and rotational invariance provides more flexibility for formation flights in constrained spaces.

\begin{figure*}
    \begin{center}  
         \includegraphics[width=1.8\columnwidth]{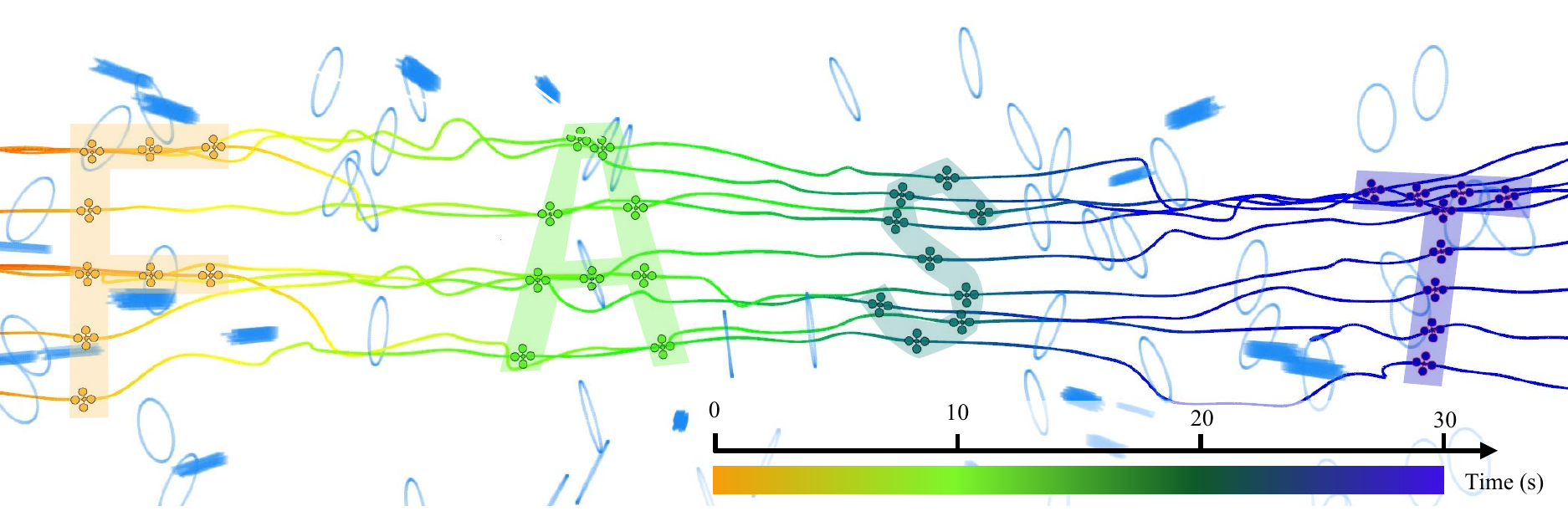}
    \end{center}
    \vspace{-0.5cm}
    \caption{Formation multiple transformation simulation. The different colors of the trajectories and robots correspond to the different timestamps. We choose four specific timestamps, and the corresponding formation shapes constitute ``FAST" (\url{http://zju-fast.com/}).}
    \label{fig:sim_fast}
\end{figure*}

\begin{figure}
    \begin{center}
         \includegraphics[width=0.9\columnwidth]{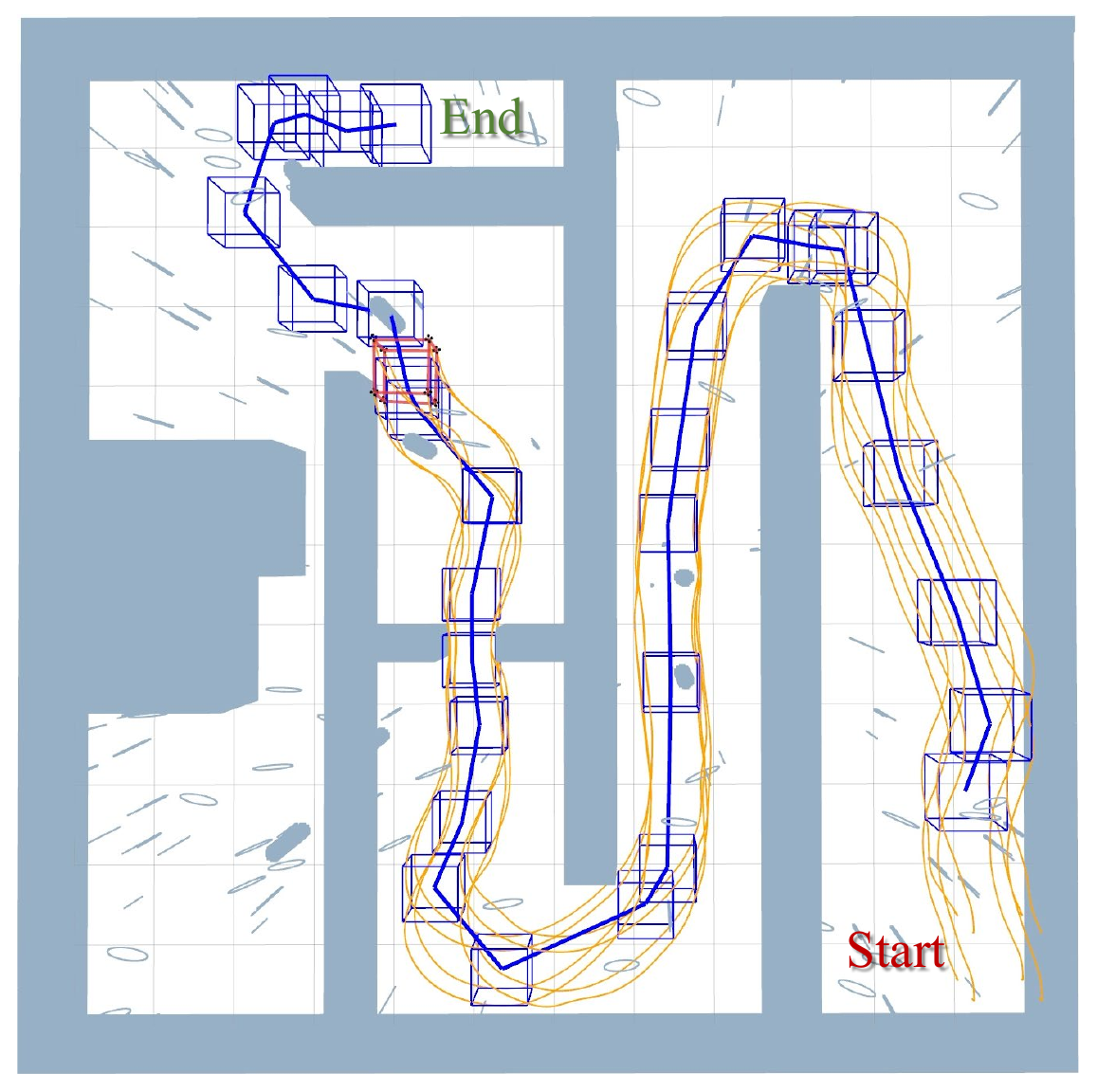}
    \end{center}
    \vspace{-0.5cm}
    \caption{Formation flight in a maze map. The formation-level global path (blue line) avoids walls while ignoring small obstacles. Then swarm formation follows the global path and generates local trajectories to avoid all obstacles and maintains the formation shape.}\label{fig:sim_maze}
    \vspace{-0.5cm}
\end{figure}

Then we design a 3-D formation shape transformation experiment to testify the reorganization ability of our method, as shown in Fig.~\ref{fig:real_8}.
In Fig.~\ref{fig:real_8}(a), the desired formation is cube shape, but the swarm robots navigate from the unconverged initial positions and improper initial task assignments.
In the beginning, the swarm robots quickly call the GRLR strategy, then make the formation shape quickly converge to the desired square shape, as shown in Fig.~\ref{fig:real_8}(b).
Then, the swarm robots receive a formation transformation command from the station laptop and converge to a double-arrow shape, as shown in Fig.~\ref{fig:real_8}(c).
The top view of the navigation process is shown in Fig.~\ref{fig:real_8}(e).
From the light blue executed trajectories, we can see that the flight behavior of swarm robots tends to be consistent during time $[4,10]$ and time $[14,20]$.
Moreover, during time $[0,4]$ and time $[10,14]$, swarm robots try to reach a swarm consensus and frequently adjust flight behavior to form the desired formation shape.
A more accurate numerical analysis can be seen from Fig.~\ref{fig:real_8}(d).
When the formation system is far from the convergence state or meets a formation transformation command, swarm robots adjust formation alignment and task assignment by calling the GRLR strategy at the time corresponding to the dotted line.
According to the similarity error represented by the red line, we can see that except for the non-convergence state at the initial moment and sudden formation transformation, the swarm formation can maintain the desired shape while avoiding obstacles.

Finally, we conduct a 16-drone swarm formation flight experiment outdoors.
To the best of our knowledge, this is the largest fully autonomous formation flight experiment in a complex outdoor environment.
As shown in Fig.~\ref{fig:main}, 16 drones form a triangular queue shape and successfully traverse an obstacle-rich area without collision.
This area has many street trees, stakes, and street lamps.
This experiment proves the robustness and large-scale ability of our proposed method.
For more details, please view the experimental video\footnote{https://www.youtube.com/watch?v=uEMyvPxYqmA}.

\begin{figure}
    \begin{center}
         \includegraphics[width=1.0\columnwidth]{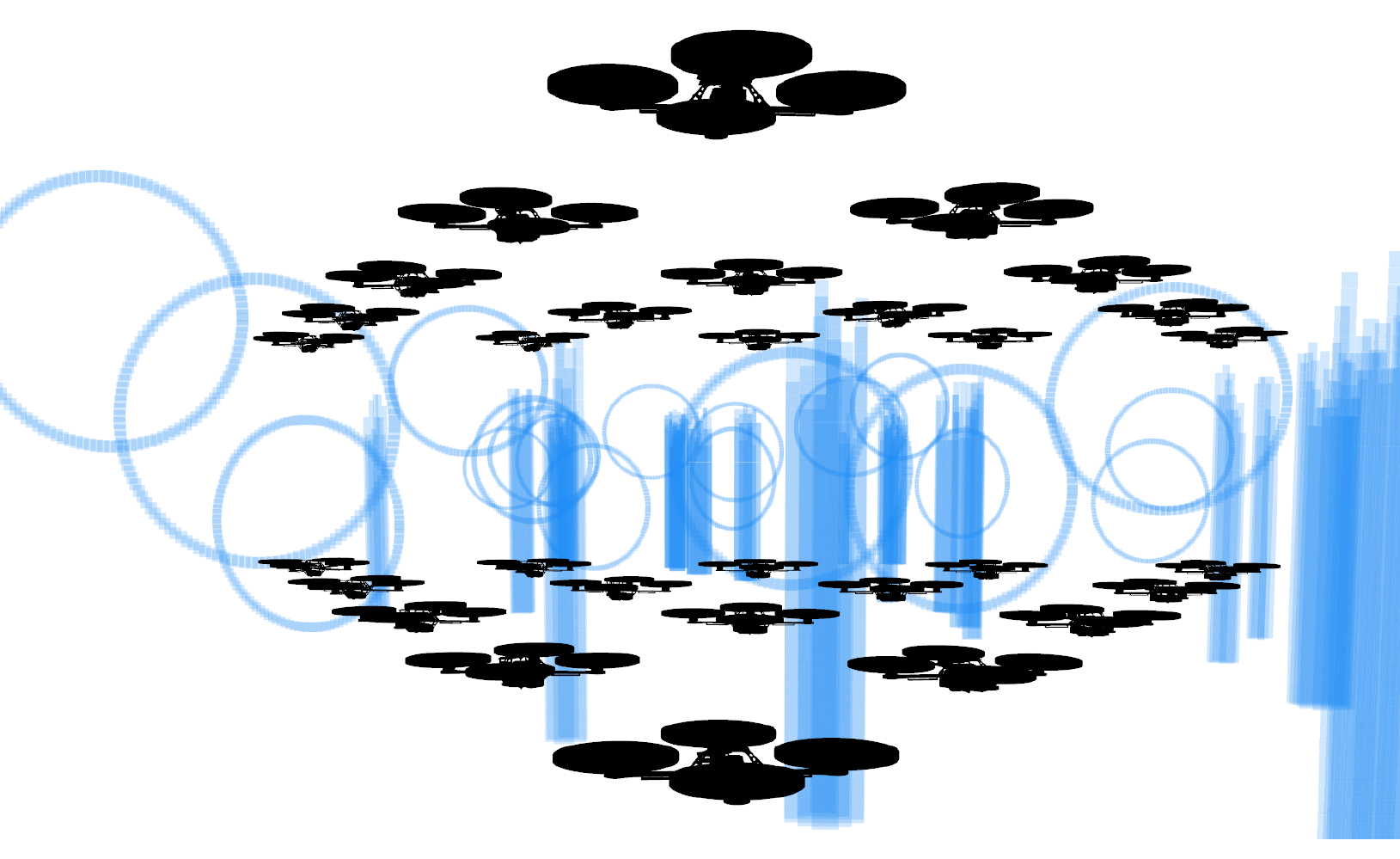}
    \end{center}
    \vspace{-0.5cm}
    \caption{Large-scale formation flight from far to near. It can be seen from the executed trajectories that the swarm robots still maintain the formation while avoiding obstacles.}\label{fig:30_swarm}
    \vspace{-0.5cm}
\end{figure}

\vspace{-0.5cm}
\subsection{Simulation Experiments}
In order to comprehensively show the characteristics of the proposed formation flight method, we also conduct several simulation experiments to supplement the real-world experiments.
All simulation experiments are run on a desktop with an Intel i7-12700 CPU in real-time.

Firstly, we design a formation multiple transformation simulation to testify swarm reorganization ability when coping with emergency changes in dense environments, as shown in Fig.~\ref{fig:sim_fast}.
The different colored trajectories represent the time flow of the formation flight from left to right.
The robots of the same color correspond to the formation shape at that timestamp.
The command of formation transformation is given in real-time instead of being set in advance. 
The swarm robots must overcome the instantaneous change of desired formation shape and quickly converge to the new formation state.
Thanks to the excellent ability of the proposed method to rapidly transform formation in complex environments, we finally generated the acronym "FAST" for our laboratory.

Then, we set up a special maze simulation environment consisting of walls and many small obstacles such as posts and rings.
In this simulation, we aim to verify swarm agreement ability.
The formation-level global path finding method first runs in the global map, which is a centralized method proposed in Sec.\ref{subsec:global-path}.
Then, it generates a global path that considers the scale of formation shape, as shown by the blue lines in Fig.~\ref{fig:sim_maze}.
The blue cube shapes represent sampling points.
After that, 8 robots form a cube formation, navigate following the global path and generate local formation trajectories using distributed methods in Sec.\ref{sec:trajectory optimization}.
This simulation demonstrates that our formation flight system can better accommodate the obstacle constraints and provide safer guidance for the formation flight.

Finally, to test our method's effectiveness with large-scale irregular formations, we design a double-arrow formation consisting of 30 drones.
As depicted in Fig.~\ref{fig:30_swarm}, the swarm successfully avoids the obstacles, and the desired formation is well preserved during the flight.

\section{Conclusion and future work}
\label{sec:conclusion}
In this paper, we analyze the core dilemmas to achieve formation flight in dense environments in detail and accurately summarize \textit{PAPER} criteria to solve the above problems.
Then we propose a hierarchical formation flight architecture composed of graph-based formation definition, distributed formation trajectory optimization, swarm reorganization method, and swarm agreement methods.
The proposed complete formation flight system satisfies all \textit{PAPER} criteria and achieves excellent performance in maintaining cooperative formation flight in dense environments.
We design comprehensive benchmarks in terms of adaptability, predictability, elasticity, resilience, and efficiency to verify the outstanding performance of our proposed method.
Finally, we conduct abundant real-world experiments and simulations to prove that we solve the problem of autonomous formation flight in dense environments within large-scale swarms.

In the future, we intend to further improve the efficiency of distributed formation flight method through local information propagation of sub-graphs.
Furthermore, while our work currently requires an operator to manually determine the formation shape, the research on optimizing formation shape is a promising area. We believe it has the potential to showcase more intelligent and cooperative swarm behavior, ultimately leading to an enhanced task capacity.

\appendix
\subsection{The Closed-form Solution of Alignment Problem}
\label{app:alignment}
Now we derive the closed-form solution to the formation alignment problem. Assume after the assignment phase, we have the current robot position $\mathbf{lg}_i$ and desired formation $\mathbf{q}_i$ with optimal matches.
We use $c_i$ to denote the $i^{th}$ term of the alignment objective in (\ref{equ:alignment problem}). 
Then we have
\begin{equation}
\begin{aligned}
     c_i &= w_i \cdot \parallel \mathbf{lg}_i - (s\,\mathbf{q}_i + \mathbf{d})\parallel^2,\\
     & = w_i\, (\mathbf{lg}_i - s\,\mathbf{q}_i - \mathbf{d})^T(\mathbf{lg}_i - s\,\mathbf{q}_k - \mathbf{d}),\\
     & = w_i\mathbf{lg}_i^T\mathbf{lg}_i + s^2\,w_i\,\mathbf{q}_i^T\mathbf{q}_i - 2s\,w_i\,\mathbf{lg}_i^T\mathbf{q}_i + \\
     & \,\,\,\,\,\,\,\,  2s\,w_i\,\mathbf{q}_i^T\mathbf{d} - 2w_i\mathbf{lg}_i^T\mathbf{d} + w_i\mathbf{d}^T\mathbf{d}.
\end{aligned}
\end{equation}
We use $F$ to denote the alignment objective in (\ref{equ:alignment problem}). Then the objective $F$ can be written as
\begin{equation}
\begin{aligned}
\label{equ:objective F}
     F(s,\mathbf{d}) & = \sum_i^n c_k, \\
                     & = b_{lg} + s^2\,b_q - 2s\,b_{lg,q} \\
                     & \,\,\,\,\,\,\,\, + 2s\,\hat{\mathbf{q}}^T\mathbf{d} -2\hat{\mathbf{lg}}^T\mathbf{d} + \mathbf{d}^T\mathbf{d},
\end{aligned}
\end{equation}
where
\begin{equation}
\begin{aligned}
    b_{lg} & = \sum_i^n w_i\,\mathbf{lg}_i^T\mathbf{lg}_i, \\
    b_q & = \sum_i^n w_i\,\mathbf{q}_i^T\mathbf{q}_i, \\
    b_{lg,q} & = \sum_i^n w_i\,\mathbf{lg}_i^T\mathbf{q}_i,
\end{aligned}
\end{equation}
\begin{equation}
    \hat{\mathbf{q}} = \sum_i^n w_i \mathbf{q}_i,\;\,\hat{\mathbf{lg}} = \sum_i^n w_i \mathbf{p}_i.
\end{equation}
Since the awareness weights $w_i$ are the outputs of a \textrm{softmax} function, we have the property $\sum_i^n w_i = 1$.
Note that we use this property to simplify the last term in (\ref{equ:objective F}). 
Now we need to prove that the objective $F$ in (\ref{equ:objective F}) is convex w.r.t. scale parameter $s$ and translation $\mathbf{d}$. The Hessian matrix of the objective is given by
\begin{equation}
    \mathbf{H} = 
    \left[\begin{array}{cc}
         \frac{\partial^2 F}{\partial s^2}& \frac{\partial^2 F}{\partial s \partial \mathbf{d}} \\
        \frac{\partial^2 F}{\partial \mathbf{d} \partial s} & \frac{\partial^2 F}{\partial \mathbf{d}^2}
    \end{array}\right]
    =
    \left[\begin{array}{cc}
         2b_q& 2\hat{\mathbf{q}}^T \\
       2\hat{\mathbf{q}} & 2\mathbf{I}_{3\times3} 
    \end{array}\right]\in \mathbb{R}^{4\times4}.
\end{equation}
The eigenvalues $\lambda$ of the Hessian are as follows
\begin{equation}
\label{equ:lambda}
    \lambda = \left\{2,\,\,2,\,\,(b_q+1) \pm \sqrt{(b_q +1)^2 -4b_{w}}\,\,\right\},
\end{equation}
where
\begin{equation}
\label{equ:bw}
\begin{aligned}
    b_w & = b_q - \hat{\mathbf{q}}^T\hat{\mathbf{q}},\\
        & = \sum_i^n w_i \,\mathbf{q}_i^T\mathbf{q}_i - (\sum_i^n w_i\,\mathbf{q}_i)^T(\sum_i^n w_i\,\mathbf{q}_i).
\end{aligned}
\end{equation}

We need to prove that (\ref{equ:bw}) is non-negative, then all the eigenvalues in (\ref{equ:lambda}) will be non-negative, thus the Hessian is semi-positive definite, the objective in (\ref{equ:objective F}) will be convex w.r.t. variables $s$ and $\mathbf{d}$.

For the non-negative weights $w_i$, we have the property $\sum_i w_i = 1$ from the \textrm{softmax} function, so we construct a convex function $h(\mathbf{x}) = \mathbf{x}^T\mathbf{x}$, and apply weighted Jensen's Inequality. We get
\begin{equation}
    w_1\,h(\mathbf{q}_1) + \cdots +w_n\,h(\mathbf{q}_n) \geq h(w_1\mathbf{q}_1+ \cdots +w_n\mathbf{q}_n),
\end{equation}
\begin{equation}
    \Rightarrow  \sum_i^n w_i \,\mathbf{q}_i^T\mathbf{q}_i \geq (\sum_i^n w_i\,\mathbf{q}_i)^T(\sum_i^n w_i\,\mathbf{q}_i).
\end{equation}
Thus (\ref{equ:bw}) is nonnegative, the objective $F$ is convex. 
Then we can obtain the closed-form solution of this alignment problem by solving
\begin{equation}
\begin{aligned}
     \left\{\begin{matrix}
     \begin{aligned}
         \partial F/\partial s &= 2s\,b_q - 2b_{lg,q} + 2\hat{\mathbf{q}}^T\mathbf{d} = 0,\\ 
         \partial F/\partial \mathbf{d} &= 2s\,\hat{\mathbf{q}} - 2\hat{\mathbf{lg}} + \mathbf{d} = 0.
     \end{aligned}
\end{matrix}\right.
\end{aligned}
\end{equation}
The close-form solution of problem (\ref{equ:alignment problem}) is given by
\begin{equation}
    \begin{aligned}
        s^{*} &= \frac{b_{lg,q} - \hat{\mathbf{lg}}^T\hat{\mathbf{q}}}{b_q - \hat{\mathbf{q}}^T\hat{\mathbf{q}}}, \\
        \mathbf{d}^* &= \hat{\mathbf{lg}} - s^*\,\hat{\mathbf{q}}.
    \end{aligned}    
\end{equation}

\bibliography{references}

\vspace{-0.5cm}
\begin{IEEEbiography}[{\includegraphics[width=1in,height=1.25in,clip,keepaspectratio]{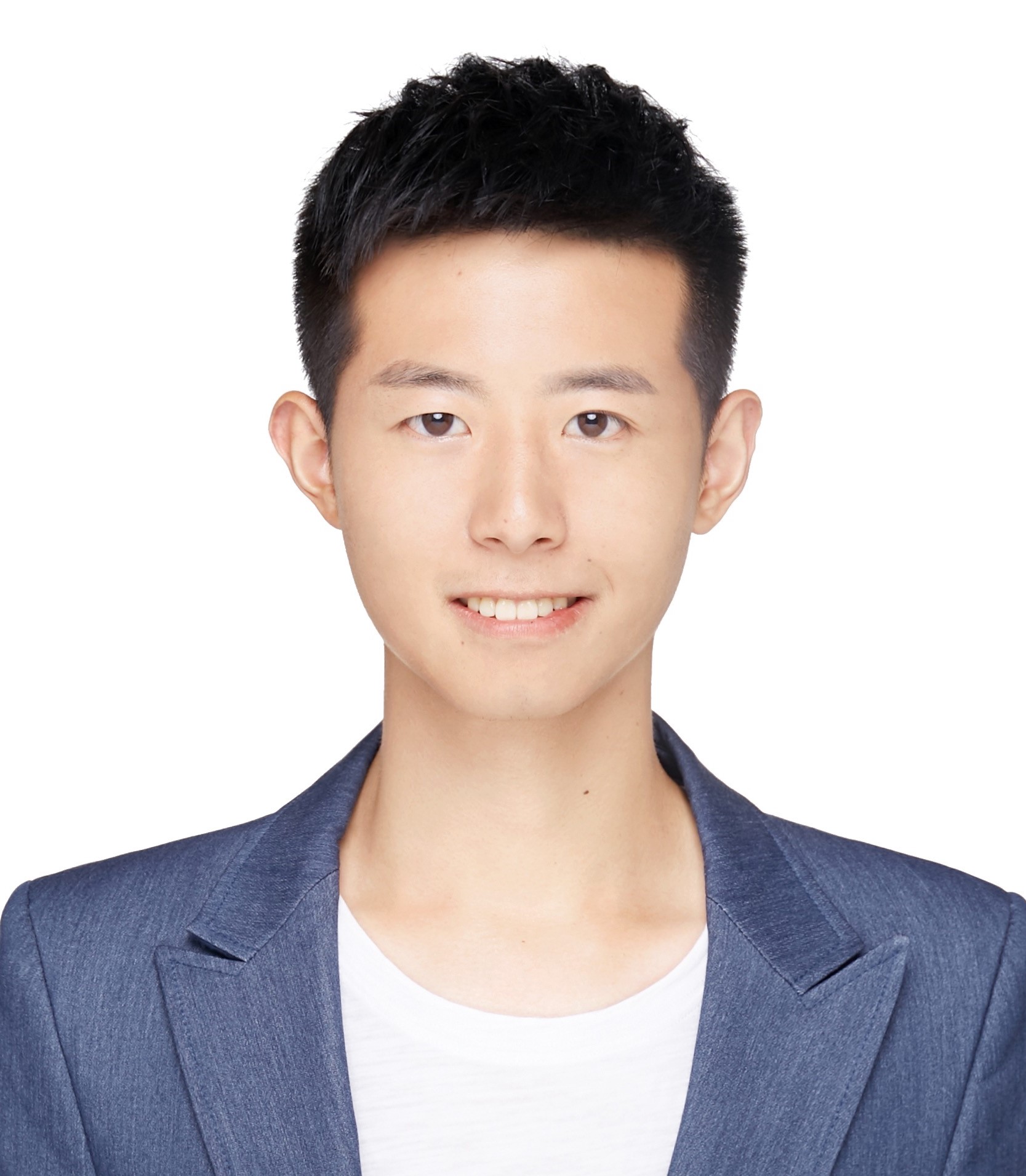}}]{Lun Quan}
received the B.Eng. degree in control science and engineering in 2019 from Zhejiang University, Hangzhou, China, where he is currently working toward a Ph.D. degree in control science
and engineering.

His research interests include motion planning for multi-robot systems, swarm intelligence, and autonomous navigation for unmanned vehicles.
\end{IEEEbiography}

\vspace{-0.5cm}
\begin{IEEEbiography}[{\includegraphics[width=1in,height=1.25in,clip,keepaspectratio]{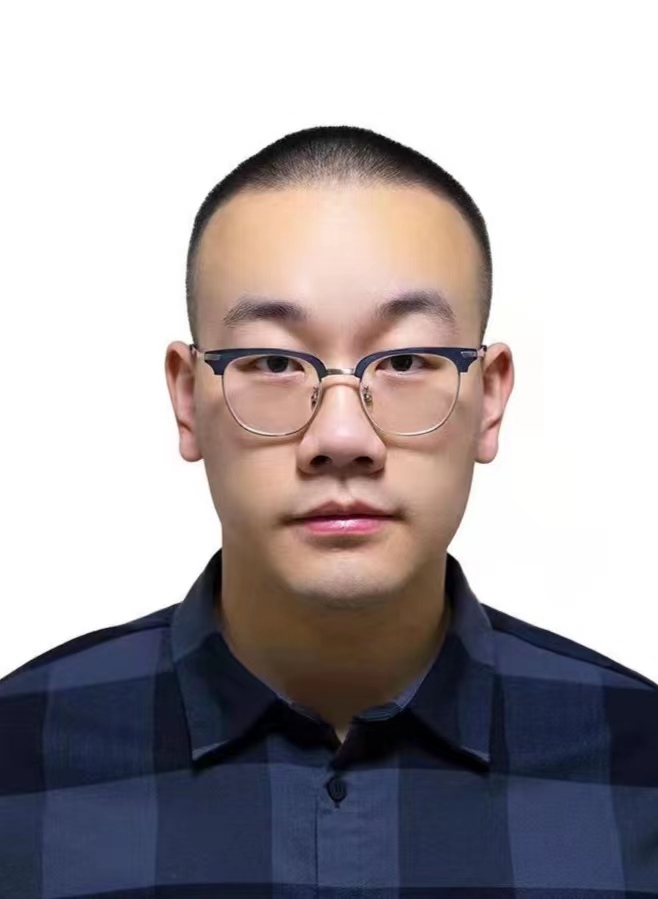}}]{Longji Yin}
received the B.Eng. degree in automation from Zhejiang University, Zhejiang, China, in 2019 and the M.Sc. degree in robotics from Johns Hopkins University, Maryland, U.S. in 2021. In 2022, he worked as a research assistant at Zhejiang University, Zhejiang, China. He is currently working toward a Ph.D. degree in mechanical engineering at the University of Hong Kong, Hong Kong. 

His research interests include motion planning and mapping for autonomous aerial robots.
\end{IEEEbiography}

\vspace{-0.5cm}
\begin{IEEEbiography}[{\includegraphics[width=1in,height=1.25in,clip,keepaspectratio]{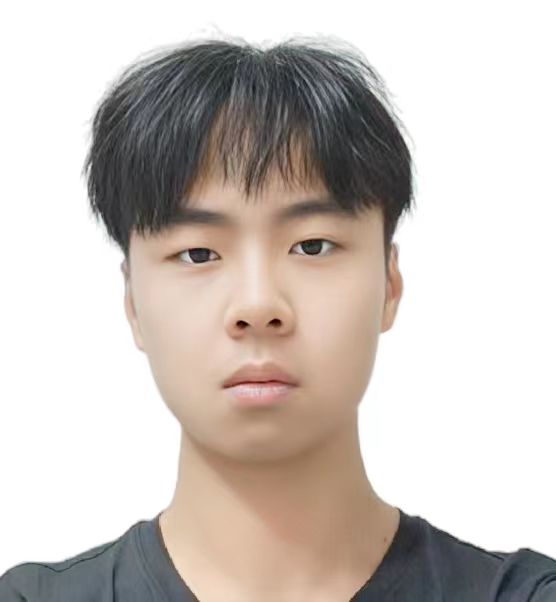}}]{Tingrui Zhang}
received the B.Eng. degree in automotive engineering from Beijing Institute of Technology, Beijing, China, in 2022. 
He is currently pursuing an M.Phil. degree in control engineering at Zhejiang University, Hangzhou, China.

His research interests include motion planning for autonomous robots and numerical optimization.
\end{IEEEbiography}

\vspace{-0.5cm}
\begin{IEEEbiography}[{\includegraphics[width=1in,height=1.25in,clip,keepaspectratio]{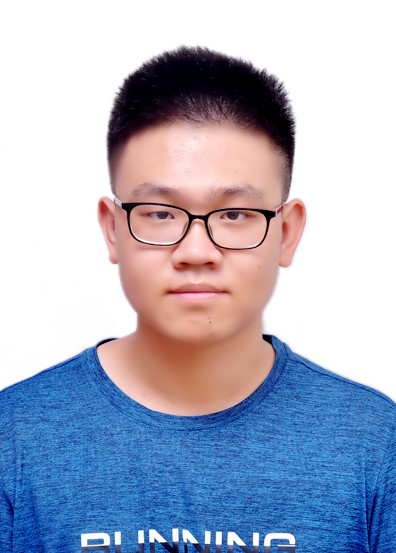}}]{Mingyang Wang}
received the B.Eng. degree in control science and engineering from Zhejiang University, Hangzhou, China, in 2022. He is currently working toward an M.Phil. degree in control science and engineering from Zhejiang University, Hangzhou, China. 

His research interests include motion planning and control for aerobatic flight.
\end{IEEEbiography}

\vspace{-0.5cm}
\begin{IEEEbiography}[{\includegraphics[width=1in,height=1.25in,clip,keepaspectratio]{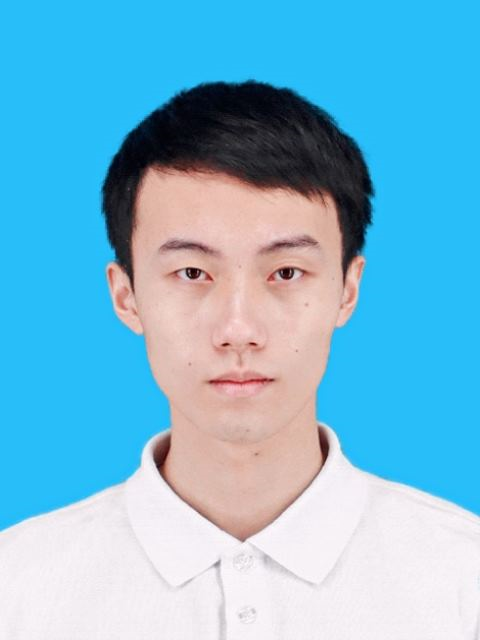}}]{Ruilin Wang}
received the B.Eng. degree in control science and engineering from Zhejiang University, Hangzhou, China, in 2023. He is currently working toward an M.Phil. degree in artificial intelligence at Sun Yat-sen University, Zhuhai, China.

His research interests include aerial robots and motion planning.
\end{IEEEbiography}

\vspace{-0.5cm}
\begin{IEEEbiography}[{\includegraphics[width=1in,height=1.25in,clip,keepaspectratio]{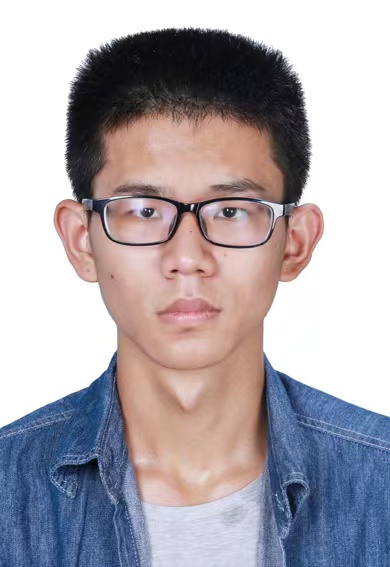}}]{Sheng Zhong}
received the B.Eng. degree in control science and engineering from Zhejiang University, China, in 2023. He is currently working toward a Ph.D. degree in control science and engineering at Hunan University, Changsha, China.

His research interests include motion estimation based on event cameras.
\end{IEEEbiography}

\vspace{-0.5cm}
\begin{IEEEbiography}[{\includegraphics[width=1in,height=1.25in,clip,keepaspectratio]{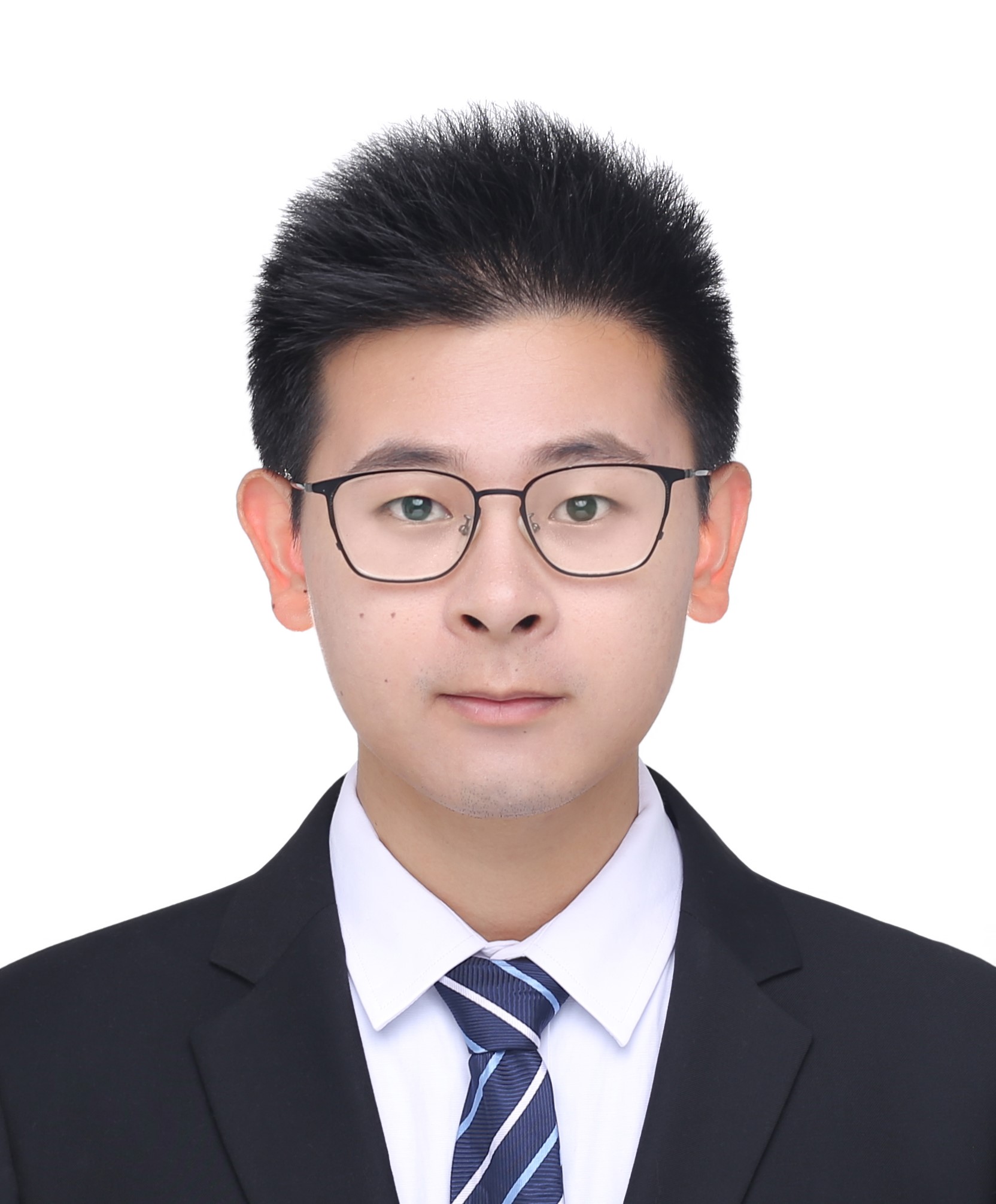}}]{Xin Zhou}
received the B.Eng. degree in electrical engineering and automation from China University of Mining and Technology, Xuzhou, China, in 2019. He is currently working toward the Ph.D. degree in control engineering from Zhejiang University, Hangzhou, China. 

His research interests include motion planning and mapping for aerial swarm robotics.
\end{IEEEbiography}

\vspace{-0.5cm}
\begin{IEEEbiography}[{\includegraphics[width=1in,height=1.25in,clip,keepaspectratio]{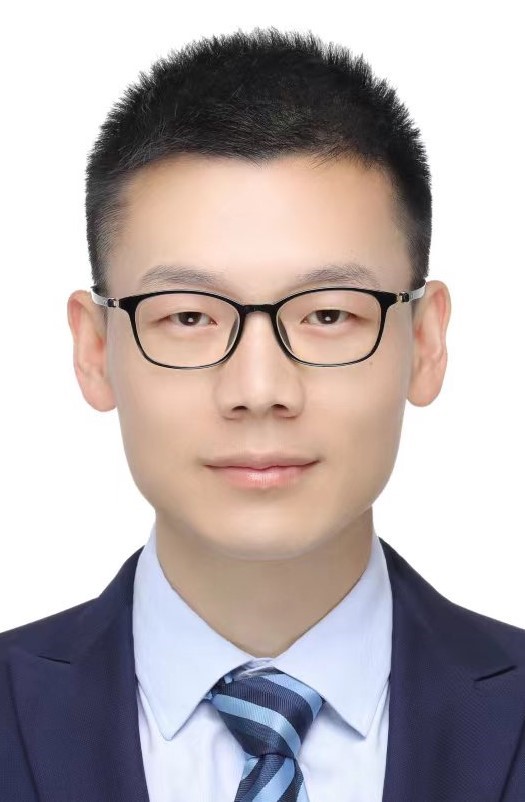}}]{Yanjun Cao}
received his Ph.D. degree in computer and software engineering from the University of Montreal, Polytechnique Montreal, Canada, in 2020.
He is currently an associate researcher at the Huzhou Institute of Zhejiang University, as a PI in the Center of Swarm Navigation. He leads the Field Intelligent Robotics Engineering (FIRE) group of the Field Autonomous System and Computing Lab (FAST Lab). 

His research focuses on key challenges in multi-robot systems, such as collaborative localization, autonomous navigation, perception, and communication.
\end{IEEEbiography}

\vspace{-0.5cm}
\begin{IEEEbiography}[{\includegraphics[width=1in,height=1.25in,clip,keepaspectratio]{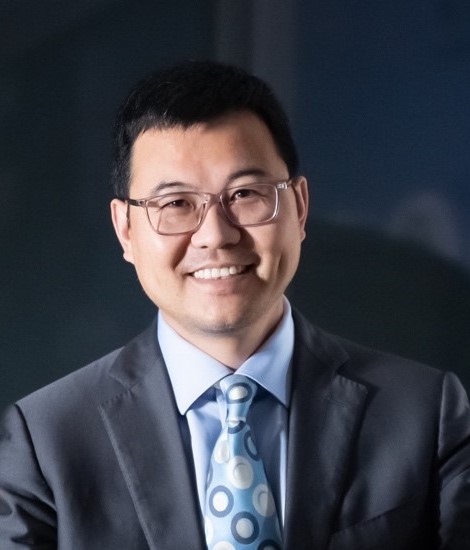}}]{Chao Xu}
received his Ph.D. in Mechanical Engineering from Lehigh University in 2010. He is currently Associate Dean and Professor at the College of Control Science and Engineering, Zhejiang University. He serves as the inaugural Dean of ZJU Huzhou Institute, as well as plays the role of the Managing Editor for \textit{IET Cyber-Systems \& Robotics}. 

His research expertise is Flying Robotics, Control-theoretic Learning. Prof. Xu has published over 100 papers in international journals, including \textit{Science Robotics} (Cover Paper), \textit{Nature Machine Intelligence} (Cover Paper), etc. Prof. Xu will join the organization committee of the IROS-2025 in Hangzhou.
\end{IEEEbiography}

\vspace{-0.5cm}
\begin{IEEEbiography}[{\includegraphics[width=1in,height=1.25in,clip,keepaspectratio]{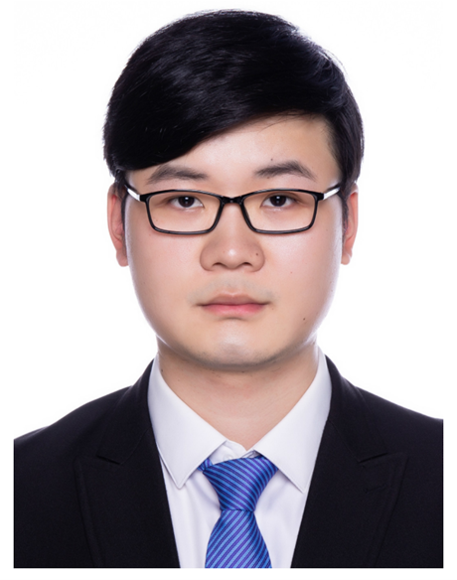}}]{Fei Gao}
received the Ph.D. degree in electronic and computer engineering from the Hong Kong University of Science and Technology, Hong Kong, in 2019.  

He is currently a tenured associate professor at the Department of Control Science and Engineering, Zhejiang University, where he leads the Flying Autonomous Robotics (FAR) group affiliated with the Field Autonomous System and Computing (FAST) Laboratory. 
His research interests include aerial robots, autonomous navigation, motion planning, optimization, and localization and mapping. 
\end{IEEEbiography}

\end{document}